\documentclass[runningheads]{llncs}
\usepackage{graphicx}
\usepackage{tabularray}
\usepackage{soul}
\usepackage{xcolor}
\usepackage{multirow}
\usepackage{hyperref}
\usepackage{tabularx}
\usepackage{array}
\usepackage[most]{tcolorbox}
\usepackage{subcaption}
\usepackage{booktabs} 
\usepackage{tikz}
\usepackage{xspace}
\usepackage{algorithm}
\usepackage{algpseudocode}
\usepackage{amsfonts}
\usepackage[numbers]{natbib}

\usepackage{enumitem}
\usepackage{wrapfig}

\usepackage[utf8]{inputenc}
\usepackage{nomencl}
\makenomenclature

\usepackage{etoolbox}
\usepackage{comment}

\newboolean{showappendix}
\setboolean{showappendix}{false}

\ifthenelse{\boolean{showappendix}}{
  \includecomment{appendixcontent}
}{
  \excludecomment{appendixcontent}
}

\makeatletter

\newcommand{\algo}{\texttt{DP-TLDM}\xspace}

\newif\ifcomm
\commfalse 
\newcommand{\lc}[1]{{\color{magenta} \ifcomm \texttt{} Lydia: #1 \fi}}

\begin{document}

\title{
\algo: Differentially Private Tabular Latent Diffusion Model}


\author{Chaoyi Zhu\inst{1}\orcidID{0000-0001-6895-7237} \and
Jiayi Tang\inst{1}\orcidID{0009-0005-4827-6406} \and 
Juan F. Pérez\inst{3}\orcidID{0000-0003-4732-1621} \and
Marten van Dijk\inst{4}\orcidID{0000-0001-9388-8050}\and
Lydia Y. Chen\inst{1,2}\thanks{Corresponding Author.}\orcidID{0000-0002-4228-6735}}

\institute{TU Delft, Delft, Netherlands \\
\email{c.zhu-2@tudelft.nl}, \email{j.tang-14@student.tudelft.nl} \and
University of Neuchâtel, Neuchâtel, Switzerland\\
\email{lydiaychen@ieee.org} \and
Universidad de los Andes, Bogotá, Colombia \\
\email{jf.perez33@uniandes.edu.co} \and
Centrum Wiskunde \& Informatica, Amsterdam, Netherlands \\
\email{marten.van.dijk@cwi.nl}
}

\maketitle
\begin{abstract}
Synthetic data from generative models emerges as the privacy-preserving data sharing solution. Such a synthetic data set shall resemble the original data without revealing identifiable private information. 
Till date, the prior focus on limited types of tabular synthesizers and small number of privacy attacks, particularly on Generative Adversarial Networks, and overlooks membership inference attacks and defense strategies, i.e., differential privacy.
Motivated by the conundrum of keeping high data quality and low privacy risk of synthetic data tables, we propose \algo, \textbf{D}ifferentially \textbf{P}rivate  \textbf{T}abular \textbf{L}atent \textbf{D}iffusion \textbf{M}odel, which is composed of an autoencoder network 
to encode the tabular data and a latent diffusion model to synthesize the latent tables. Following the emerging $f$-DP framework, we apply DP-SGD to train the auto-encoder in combination with batch clipping and use the separation value as the privacy metric to better capture the privacy gain from DP algorithms. Our empirical evaluation demonstrates that \algo is capable of achieving a meaningful theoretical privacy guarantee while also significantly enhancing the utility of synthetic data. Specifically, compared to other DP-protected tabular generative models, \algo improves the synthetic quality by an average of 35\% in data resemblance, 15\% in the utility for downstream tasks, and 50\% in data discriminability, all while preserving a comparable level of privacy risk.  







\end{abstract}

\keywords{synthetic tabular data, deep generative models, differential privacy}

\section{Introduction}
\label{sec:intro}

High-quality synthetic data obtained from generative models are increasingly used to augment and substitute real data, boosting data utility for individuals and enterprises~\citep{ali2022spot, chambon2022roentgen, chambon2022adapting}. 
As synthetic data resembles real data, it can be used to accelerate data-driven knowledge discovery and still abide by data protection regulations, e.g., GDPR~\citep{eu2016gdpr}, which restricts the collection and accessibility of real data. 
A key requirement for the adoption of these models in the industry is their ability to preserve the privacy of the real data~\citep{yoon2020anonymization, giomi2023unified}. 
Consider for instance medical institutes that own a subset of patients' data that cannot be shared freely and are subject to lengthy regulatory auditing. Alternatively, through a trusted party that first trains the generative model, a complete set of patients' synthetic data can be generated and distributed to all institutes that in turn design their own medical analysis based on these data. While the focus of generative models lies on producing synthetic data highly similar to and indiscernible from the real data, a rising concern is the real data privacy leakage caused by the synthetic data~\citep{carlini2023extracting, hu2023membership, duan2023diffusion}. 
These studies highlight privacy vulnerabilities associated with synthetic data across specific domains, especially in image processing, and pertain to various generative models, including Bayesian networks, generative adversarial networks (GANs), and, more recently, diffusion processes. These privacy risks are materialized in attacks that are able to obtain training data, 
under various assumptions on the availability of model knowledge, i.e., white-box v.s.~black-box attacks. 

In the tabular data domain, Anonymeter~\citep{giomi2023unified} is the first framework that focuses on the privacy and utility trade-off of synthetic tables and introduces three privacy attacks relevant to tabular data: singling out attacks, linkability attacks, and attribute inference attacks.  
While it sheds light on quantifying the privacy-utility trade-off,~\citep{giomi2023unified} focuses on tabular GAN models, i.e., CTGANs~\cite{xu2019modeling}, leaving the question of how this tradeoff behaves for different tabular generative models unaddressed. 
More importantly, the critically important category of Membership Inference Attacks (MIA)~\citep{sompalli2019logan,chen2020gan2023trojdiff,mukherjee2021privgan}, which present stronger adversarial assumptions and infer whether specific data records are present in the training set, is overlooked. 
Last but not least, the impact of adopting privacy-enhancing strategies, such as differential privacy, on synthetic tabular data is largely unexplored 
by prior art.

Differential privacy (DP)~\citep{dwork2006calibrating} has received much attention as a solution to the problem of preserving individual privacy when releasing data.
To incorporate DP in the training of deep neural models with stochastic gradient descent (SGD), DP-SGD~\citep{abadi2016deep} 
obfuscates gradient updates 
by adding calibrated statistical noise that is controlled by a privacy budget. There are two main DP analysis frameworks, ($\epsilon,\delta$)-DP~\citep{dwork2006calibrating}, and emerging f-DP~\citep{dong2021gaussian}, where the former uses $\epsilon$ to define the privacy budget and the latter uses the separation value, which is the distance between the actual trade-off function of false positive and false negative and the ideal one, where no privacy leaks. 
A smaller privacy budget or separation value 
leads to adding more obfuscation noise to the gradients, degrading the performance of the underlying models. It is a long-standing challenge to apply meaningful $\epsilon$ or separation value while achieving satisfactory learning outcomes for image classification~\citep{abadi2016deep} and synthesizing~\citep{jordon2018pate}.  
The privacy enhancement of DP on tabular generative models is yet to be explored, especially concerning different genres of generative models.

Based on the insights from our empirical study, we propose a novel differentially private latent tabular diffusion model, \algo, composed of an autoencoder and a diffusion model. Different from the existing TabDDPM, we first encode the tabular data into a continuous latent space, using the autoencoder network. This brings the advantage of a unified and compact representation of categorical variables, in contrast to the typical one-hot encoding. We then use the latent representation as input to the backbone diffusion model, which captures the data synthesis as a sequence of denoising processes~\citep{kotelnikov2023tabddpm}. 
To guard the proposed latent tabular diffusion against privacy attacks, we  train the auto-encoder using DP-SGD. We follow the $f$-DP framework~\citep{dong2021gaussian}, which provides a precise parameterization of DP-SGD, specifically through the separation measure --- a metric quantifying the maximum difference between false positive and false negatives when comparing a random guess and DP-protected algorithm. 
Thanks to the post-processing guarantees of DP, the backbone latent diffusion training is also protected by the DP. 
We extensively evaluate the proposed $\algo$ against DP-CTGAN and DP-TabDDPM, where DP-SGD is used to train CTGAN and TabDDPM, showing a remarkable performance --- reducing the privacy risk especially against MIA while maintaining a significant high data utility compared to the other two synthesizers. 
We make the following concrete contributions: 
\begin{itemize}[leftmargin=0.2cm]
\item  We design \algo, a novel latent tabular diffusion model trained by DP-SGD that uses batch clipping on gradients and Gaussian noising mechanism. Our approach leverages the $f$-DP framework, where we propose a new theoretical privacy metric, termed \textit{separation}, to enhance privacy guarantees.
\item Our evaluation of \algo against DP-CTGAN and DP-TabDDPM shows that \algo can effectively reduce the privacy risks while maintaining high synthetic data quality across all privacy budget values. 
As a result \algo displays similar privacy risks levels than other synthesizers, but outperforms them by an average of 35\% in data resemblance, 15\% in the utility for downstream tasks, and a 50\% in data discriminability. 
\end{itemize}

\section{Related Studies}
In this section, we provide a general overview of the generative models and privacy on the tabular data.


\textbf{Tabular generative models} 
Current state-of-the-art introduces several {deep generative models for tabular data synthesis. TableGAN~\citep{tablegan:park:vldp18} implements an auxiliary classification model along with discriminator training to enhance column dependency in the synthetic data. CTGAN~\citep{xu2019modeling} improves data synthesis by introducing several preprocessing steps for categorical, continuous or mixed data types which encode data columns into suitable form for GAN training. The conditional vector designed by CTGAN~\citep{xu2019modeling} and later improved by Ctab-GAN~\citep{zhao2021ctab} also helps the GAN training to reduce mode-collapse on minority categories. 
Drawback of these methods is also clear that there is loss of information during the transformation from table to latent vector. Therefore, GAN cannot learn the knowledge from the information that loses during this compression. TabDDPM~\citep{kotelnikov2023tabddpm} is based on denoising diffusion probabilistic models (DDPM)~\citep{ho2020denoising}, it uses two different diffusion models to synthesize categorical and continuous columns.
\textbf{Privacy attacks} Despite the impressive performance and application of deep generative models, recent works have also raised significant concerns regarding the potential privacy risks of these models. 
A vast body of related studies on \textbf{privacy attacks} can be categorized by various attack types, including (i) membership inference attacks (MIA) \citep{shokri2017membership, carlini2023extracting, chen2020gan2023trojdiff, sompalli2019logan}, inferring whether a certain data record is in the training set; (ii) attribute inference attacks~\citep{giomidler2020synthetic, giomi2023unified}, deducing sensitive attributes of the training data; (iii) replication attacks~\citep{carlini2023extracting, hu2021model, somepalli2023diffusion}, reproducing the training data or hidden generative models; (iv) adversarial attacks~\citep{szegedy2014intriguing, milliere2022adversarial, gao2023evaluating}, deceiving generative models through crafted input data at the inference stage. \textbf{Membership inference attacks} can be further categorized into white-box, no-box and black-box attacks based on the availability of model information. In white-box attacks, where attackers have access to the internals of generators, several works \citep{carlini2023extracting, hu2023membership, zhu2023data} have proposed loss-based techniques for conducting MIA on diffusion models.  Black-box setting assumes  the prior knowledge of attackers is limited only to generated samples~\citep{zhu2023data, wu2022membership}.  

%

\textbf{Privacy enhancing methodologies} have been studied to address potential privacy risks.
{DP-SGD} and its variants have been widely adopted for privately training deep generative models. DPGAN \citep{xie2018differentially} applies the DP-SGD algorithm directly to the discriminator component within GANs. In contrast, GS-WGAN \citep{chen2020gs} implements DP-SGD on the gradients transferred from the discriminator to the generator. 
The utility of DP-SGD is also extended beyond GANs and applied in normalizing flows for tabular data synthesis \citep{waites2021differentially, lee2022differentially}.
Moreover, in the context of emerging diffusion models, adaptions of DP-SGD are considered as well. 
One study \citep{dockhorn2022differentially} applied the classic DP-SGD algorithm with one modification involving sampling multiple time steps of a single data point when computing the loss. 
Building on this, another study \citep{ghalebikesabi2023differentially} further presented the effectiveness of three other techniques, namely pre-training, augmentation multiplicity, and modified time step sampling. 
While DP-SGD is deemed a strong countermeasure for privacy leaks, it comes at the cost of sample quality and longer training times.

\section{Empirical Analysis}
\label{sec:EmpiricalAnalysis}
In this section, we put our risk-utility quantification framework described in Appendix B.3 to the test on publicly available datasets that have been extensively employed in tabular data analysis and synthesis.

\subsection{Datasets} 
We employ four 
datasets, two small (up to 20000 samples) and two larger.  
Since small datasets usually make models prone to overfitting, by comparing these datasets, we can understand how dataset size and overfitting affect the quality and privacy of synthetic data. 
Some characteristics of the datasets are listed in Table 4 in the appendix.

The \textbf{Loan dataset}~\citep{misc_loan} contains demographic information on 5000 customers. It holds 14 features divided into 4 different measurement categories, including  binary, interval, ordinal, and nominal features.  
The \textbf{Housing dataset}~\citep{misc_housing} relates to houses in a given California district and provides summary statistics based on the 1990 Census data. It comprises 20,640 instances with 1 categorical and 9 numerical features and a total of 207 missing values. The \textbf{Adult dataset}~\citep{misc_adult_2} contains information on individuals' annual incomes and related variables. It consists of 48842 instances with 14 mixed datatype features in total, and a total of 6465 missing values.  
The \textbf{Cardiovascular Heart Disease dataset}~\citep{misc_cardio} contains detailed information on the risk factors for cardiovascular disease, including 70000 instances with 13 mixed-type columns. 

For all datasets, each synthesizer generated a synthetic dataset with the same size as the training dataset for evaluation. For the privacy evaluation, 1000 records are randomly sampled from each training set for every attack.

\subsection{Privacy-utility Trade-off}
\label{sec:PU-tradeoff}


Table~\ref{tab:no_dp} 
presents detailed results quantifying both utility and risk aspects of synthetic data for all four datasets employing the five 
generative models described in Appendix B.1.  
We present the three utility metrics discussed before (i.e., resemblance, discriminability, and utility), where a higher score indicates better performance, as well as the privacy risk for the four attacks considered (Singling out, Linkability, AIA, MIA), where a lower risk indicates better performance. Due to space reasons, we keep the detailed statistics of the five MIA attacks in the appendix. 
\begin{table}[htb]
\centering
\resizebox{0.6\textwidth}{!}{%
\begin{tabular}{c|c|ccc|ccccc}
\toprule[0.9pt]
\multirow{2}{*}{Dataset} & \multirow{2}{*}{Method} & \multicolumn{3}{c|}{Quality Score $\uparrow$}   & \multicolumn{4}{c}{Privacy Risk $\downarrow$} \\ \cline{3-9} 
 &     & Resem. & Discri. & Utility  & S-out & Link & AIA   & MIA   \\ \bottomrule[0.9pt]
\multirow{6}{*}{Loan}       & CopulaGAN & 92          & 95               & 70      & 52.81        & 2.31        & 6.98  & 2.86  \\ \cline{2-9} 
                            & CTGAN     & 92          & 85               & 93      & 54.90        & 0.00        & 8.11  & 2.86  \\ \cline{2-9} 
                            & ADS-GAN   & 93          & 95               & 73      & 17.23        & 0.00        & 0.00  & 22.86 \\ \cline{2-9} 
                            & GC        & 86          & 82               & 78      & 28.68        & 0.00        & 0.00  & 5.72  \\ \cline{2-9} 
                            & TabDDPM   & 98          & 100              & 97      & 26.31        & 2.23        & 16.68 & 45.72 \\ \bottomrule[0.9pt]
\multirow{6}{*}{Housing}    & CopulaGAN & 94          & 90               & 62      & 8.14         & 0.00        & 1.54  & 20.00 \\ \cline{2-9} 
                            & CTGAN     & 94          & 92               & 64      & 12.55        & 0.45        & 0.00  & 20.00 \\ \cline{2-9} 
                            & ADS-GAN   & 93          & 87               & 74      & 1.73         & 1.43        & 0.98  & 48.58 \\ \cline{2-9} 
                            & GC        & 91          & 84               & 32      & 4.16         & 0.00        & 0.00  & 5.72  \\ \cline{2-9} 
                            & TabDDPM   & 96          & 98               & 93      & 1.30         & 0.16        & 0.00  & 88.58 \\ \bottomrule[0.9pt]
\multirow{6}{*}{Adult}      & CopulaGAN & 93          & 97               & 81      & 10.25        & 0.05        & 5.08  & 17.14 \\ \cline{2-9} 
                            & CTGAN     & 90          & 79               & 83      & 20.18        & 0.55        & 3.16  & 10.00 \\ \cline{2-9} 
                            & ADS-GAN   & 88          & 59               & 83      & 19.74        & 0.00        & 0.00  & 20.00 \\ \cline{2-9} 
                            & GC        & 80          & 50               & 56      & 32.80        & 0.38        & 2.86  & 8.58  \\ \cline{2-9} 
                            & TabDDPM   & 96          & 98               & 98      & 22.72        & 0.46        & 0.00  & 94.28 \\ \bottomrule[0.9pt]
\multirow{6}{*}{Cardio}     & CopulaGAN & 87          & 93               & 96      & 66.54        & 1.13        & 28.75 & 0.00  \\ \cline{2-9} 
                            & CTGAN     & 84          & 68               & 97      & 62.04        & 1.11        & 24.07 & 11.42 \\ \cline{2-9} 
                            & ADS-GAN   & 90          & 71               & 100     & 59.76        & 0.89        & 15.21 & 31.42 \\ \cline{2-9} 
                            & GC        & 81          & 63               & 86      & 61.02        & 0.44        & 6.30  & 22.86 \\ \cline{2-9} 
                            & TabDDPM   & 95          & 99               & 100     & 60.77        & 1.31        & 23.08 & 94.28 \\ \bottomrule[0.9pt]
\end{tabular}%
}
\caption{Quantification of Risk-Utility for Five Generative Models Across Various Datasets. Here, ``Resem." stands for Resemblance,``Distrim." refers to Discriminability, ``S-out" denotes singling out attacks, and "Link" represents linkability attacks.}
\label{tab:no_dp}
\end{table}
\textbf{Comparing the synthesizers},  
TabDDPM generates synthetic data of the highest quality, outperforming other synthesizers. Across all four datasets, TabDDPM consistently secures top-three rankings in terms of resemblance, discriminability, and utility. 
CopulaGAN displays very good results in resemblance and discriminability but scores relatively low in utility. The Gaussian Copula sits at the other end of the spectrum, being outperformed by the other synthesizers across all datasets. 

Despite the excellent performance of TabDDPM in generating high-quality synthetic data, it presents the highest risk, particularly in relation to Linkability and MIA. 
Its risk is especially high in terms of MIA attacks, where it displays a significantly higher risk than the other synthesizers. 

On the contrary, the GAN family and Gaussian Copula, while not achieving superior synthetic data quality, showcase greater resilience to Linkability, AIA, and MIA attacks. 
This suggests that:\\
\noindent\vspace{.2cm}
\noindent\fcolorbox{black}{gray!10}{\begin{minipage}{\linewidth
}
Synthetic data with higher quality tend to closely resemble the original data, potentially resulting in heightened exposure of the genuine data and increased susceptibility to exploitation by attackers, especially shown in TabDDPM.
\end{minipage}}

\textbf{Across all types of attacks}, AIA and MIA  consistently display greater efficacy, as evidenced by their higher average risk observed across the four datasets.  
Notably, Linkability, AIA, and MIA attacks consistently manifest more detrimental effects on synthesizers that demonstrate superior utility, such as TabDDPM and ADS-GAN. Conversely, the Singling Out attack emerges as the predominant threat to synthesizers with lower utility, as exemplified by Gaussian Copula and Copula GAN. 

This divergence underscores the intricate vulnerabilities of synthesizers to distinct attack methodologies. While Linkability, AIA, and MIA generally rely on the comprehensive attributes of synthetic data, the Singling Out Attack is based upon identifying outlier values within the synthetic dataset. This suggests that: \\
\noindent\vspace{.2cm}
\noindent\fcolorbox{black}{gray!10}{\begin{minipage}{\linewidth
}
Synthetic data of suboptimal quality may disclose more information about outliers to potential attackers as in Singling Out attacks. Conversely, high-quality synthetic data are prone to reveal more comprehensive and overall information of the original data as shown in Linkability, AIA and MIA attacks.
\end{minipage}}

\textbf{Regarding MIA strategies}, notable effectiveness is achieved by the NaiveGroundhog (NG), HistGroundhog (HG), and Closest Distance-Hamming (CD-H) strategies, which are able to reach success rates of 60\% or higher in some cases. These results are detailed in Table 5 in Appendix D.1.  
Remarkably, HistGroundhog consistently outperforms other MIA strategies when applied to the TabDDPM synthesizer. In contrast, the NaiveGroundhog and Closest Distance-Hamming strategies demonstrate better efficacy when employed on other synthesizers.

In contrast, Closest Distance-L2 (CD-L) and Kernel Estimator (KE) strategies, exhibit a comparatively lower level of effectiveness. Given that half of the target records for MIA are from the training data, and both strategies consistently attain success rates close to 50\%, the performance of these two strategies closely aligns with random guessing. 
This observation underscores the nuanced variations in the efficacy of MIA strategies for different synthesizer models. It indicates that:\\
\noindent\vspace{.2cm}
\noindent\fcolorbox{black}{gray!10}{\begin{minipage}{\linewidth}
Sophisticated shadow modeling approaches (HistGroundhog) exhibit heightened effectiveness when applied to high-quality synthetic data. In contrast, simpler shadow modeling methods (NaiveGroundhog) and distance-based strategies (Closest Distance-Hamming) may prove more effective when the synthetic data quality is suboptimal. 
\end{minipage}}

\textbf{Across all data sets}, 
the Linkability attack demonstrates higher average privacy risk, particularly when applied to smaller datasets such as Loan and Housing. As for other attacks, trends related to different dataset sizes are less evident. 

In terms of synthetic data utility, larger datasets (Adult and Cardio) exhibit, on average, lower resemblance and discriminability scores compared to smaller ones (Loan and Housing). 
These findings prompt that larger datasets pose more challenges to the synthesizers, as increased dataset sizes may introduce greater diversity and complexity, thereby making data synthesis more difficult. 

However, the utility scores are higher when dataset sizes increase. This phenomenon may be attributed to the fact that the utility metric is measured on the performance of downstream machine learning tasks, which are inherently influenced by the size of training data. In our experiments, the synthetic dataset size remains the same as the corresponding real dataset. Consequently, small real datasets result in small synthetic datasets, which may potentially engender suboptimal performance in machine learning tasks and lower utility scores. 

This leads us to conclude that in our experiments:\\
\noindent\vspace{.2cm}
\noindent\fcolorbox{black}{gray!10}{\begin{minipage}{\linewidth
}
The larger datasets are more challenging with regard to the data synthesis task and potentially less vulnerable to adversarial privacy attacks.
\end{minipage}}

\section{\algo}

In this section, we introduce our latent tabular diffusion model (\algo), which effectively incorporates robust privacy protections by integrating Differential Privacy (DP) techniques. Illustrated in Figure~\ref{fig:LatentDiff}, our model consists of two components: the Autoencoder and the Latent Diffusion Model. Initially, the autoencoder performs the task of encoding both continuous and categorical features in the original tabular data into a unified latent space, meanwhile ensuring DP protection is applied throughout this transformation. Subsequently, the Latent Diffusion Model conducts a Gaussian diffusion process within the latent space.
\begin{figure}[tb]
    \centering
  \includegraphics[width=.7\textwidth]{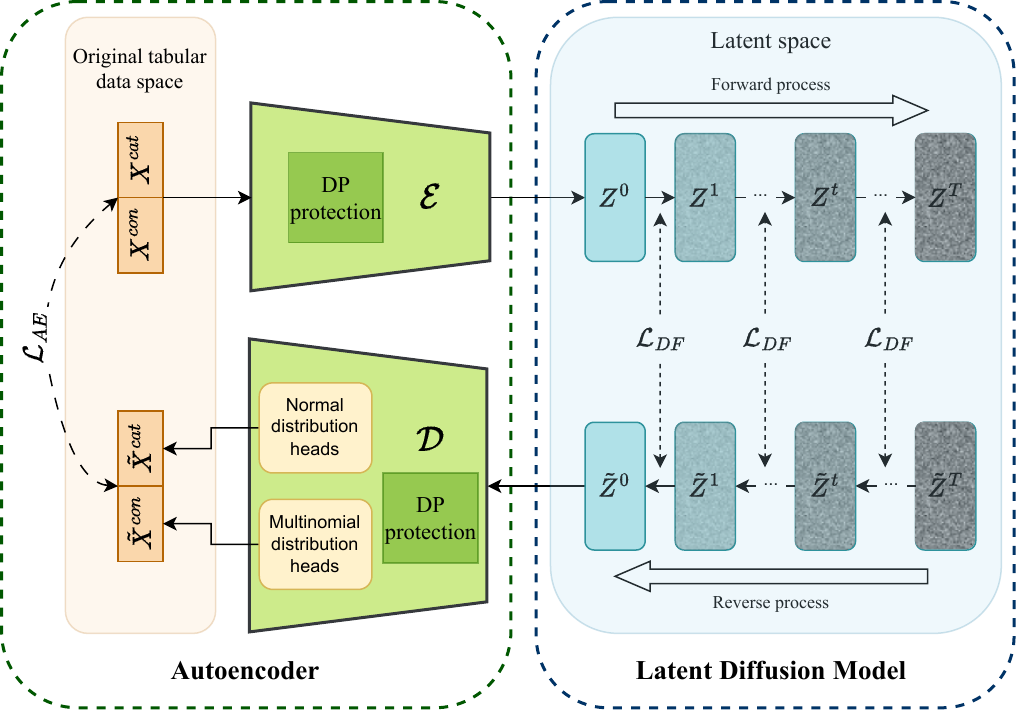}
     \caption{The latent tabular diffusion model. Given the original tabular data with both continuous and categorical features, the autoencoder first encodes both features into a cohesive latent space, with the protection of Differential Privacy (DP). The Latent Diffusion Model then executes a Gaussian diffusion process within the latent space.}
    \label{fig:LatentDiff}
\end{figure}

The essential background of diffusion models for tabular data is introduced in Section \ref{sec:DiffBackground}. Following this, in Section \ref{sec:LatentDiff}, we delineate the motivation behind the development of our latent tabular diffusion model, as well as details regarding the two components. Furthermore, the inclusion of DP-enhanced training and Differential Privacy measures are presented in Sections~\ref{sec:DPsgd} and~\ref{sec:DP}, respectively.

\subsection{Diffusion Primer}
\label{sec:DiffBackground}
 Diffusion models work with a forward process perturbing the data into Gaussian noise and a reverse process learning to recover the data from the pure noise.

Typically, given the original data $x_0$, and a total of $T$ steps, the forward process $q(x_t | x_{t-1})$ at step $t$ is modeled as a Markov chain that adds pure noise to the data. The whole forward process eventually ends at a simple distribution (e.g., standard Gaussian distribution) $p(x_T)$. The reverse process, starting at $p(x_T)$, is another Markov Chain with learned transitions $p_\theta (x_{t-1} | x_t)$, which are unknown and estimated by a neural network. 




In the realm of modeling tabular data, the inherent heterogeneity among features necessitates tailored approaches for accurate modeling. TabDDPM~\citep{kotelnikov2023tabddpm} addresses this challenge by adopting different methods for noising and denoising continuous and categorical features, as shown in Figure \ref{fig:TwoDiff}. 
TabDDPM employs Gaussian diffusion 
following
~\citep{ho2020denoising}, where the forward process gradually adds Gaussian noise to the input data, which eventually ends at $p_{con}(x_T)=\mathcal{N}(x_T;\mathbf{0},\mathbf{I})$. Conversely, in the reverse process, a neural network is trained to predict the added noise, thereby facilitating the denoising of the data. 

\begin{figure}[htbp]
    \centering
    \includegraphics[width=0.7\textwidth]{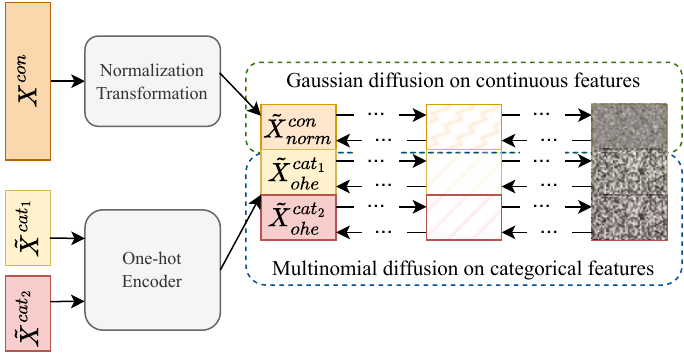}
    \caption{The architecture of TabDDPM where different methods are adopted for continuous and discrete features separately. Continuous features 
    are handled by the Gaussian diffusion process, whereas categorical features are one-hot encoded and diffused using the Multinomial diffusion process. 
    }
    \label{fig:TwoDiff}
\end{figure}

Meanwhile, categorical features are handled using Multinomial diffusion, as proposed by \citep{hoogeboom2021argmax}, with each categorical feature having a distinct Multinomial diffusion process. For a categorical feature with $K$ classes, during the forward process, uniform noise over the $K$ classes is applied to corrupt the one-hot encoded categorical feature, ending at the categorical distribution $p_{dis}(x_T)=\mathcal{C}(x_T;1 / K)$. Subsequently, the reverse process leverages a neural network to predict the probability vector to 
recover the noised data. 

TabDDPM employs one multi-layer neural network for both the Gaussian diffusion and the Multinomial diffusions. The input to the network is the concatenated representation of both the normalized continuous features and one-hot encoded categorical features. The output has the same dimensionality as the input, with the first few coordinates being the predicted Gaussian noise and the rest being the predictions of probability vectors. The model is trained by minimizing a sum of the mean-squared error for the Gaussian diffusion and the KL divergences for each multinomial diffusion.

\subsection{Tabular Latent Diffusion Model (TLDM)}
\label{sec:LatentDiff}
While different diffusion processes for continuous and categorical features in TabDDPM underscore a strategy to accommodate the diverse nature of tabular data, there are two potential drawbacks. First, the utilization of one-hot encoded representations for categorical 
columns in tabular data 
introduces significant complexity. 
For instance, in the Adult dataset, the Educational Level column consists of 16 distinct categories, resulting in the transformation of a single column into a one-hot encoded feature vector with a dimensionality of 16. Second, the separation and discrepancy in the diffusion processes for continuous and categorical features could also lead to a potential loss of inter-feature relationships and dependencies. By treating continuous and categorical features with independent diffusion processes, the model may overlook the intricate correlations that could exist between features. 

To address the mentioned limitations, we propose the latent tabular diffusion model following~\citep{rombach2022high}.  In our latent tabular diffusion model, both the continuous and categorical features are transferred to a unified continuous latent space by training an autoencoder. 
Subsequently, a unified diffusion model is deployed to noise and denoise the continuous latent features. The decoder component of the autoencoder is then employed to convert the denoised latent representation back to the original features. 

Consequently, our model mitigates the sparsity and dimensional complexity associated with the one-hot encoding technique used in TabDDPM. With a unified continuous latent space, the diffusion model benefits from a more compact and streamlined input structure. Besides, jointly embedding both types of features into a latent representation also facilitates the preservation of inter-feature correlations within the original data.

Incorporating the autoencoder component provides an additional benefit in safeguarding the model through differential privacy (DP). In DP, a limited number of training epochs is set for a certain privacy budget. 
By decoupling the training procedures of the autoencoder and diffusion model components in our model, we are able to introduce DP mechanisms specifically to the training phase of the autoencoder. Thereby, only the autoencoder component will undergo a reduction in training epochs while the diffusion model can still be sufficiently trained. This deliberate separation of training procedures effectively balances privacy preservation and model efficacy for generating tabular data.

\subsubsection{Autoencoder}
The autoencoder component in our model comprises two parts: the encoder $\mathcal{E}$ and the decoder $\mathcal{D}$. Initially, given the original tabular data $X$, containing both continuous and categorical features, the encoder $\mathcal{E}$ jointly transforms the entire $X$ into a continuous latent representation $Z = \mathcal{E}(X)$. Subsequently, the decoder $\mathcal{D}$ reconstructs the latent representation $Z$ back into the original data space, yielding $\tilde{X} = \mathcal{D}(Z)$.

To handle the heterogeneity of features in tabular data, rather than treating continuous and categorical features with separate diffusion processes like TabDDPM, we add distinct heads in the output layer of the decoder to map each feature to a probability distribution. 

For continuous features $X^{con}$, the Gaussian distribution is chosen and the head outputs the mean and variance of the distribution, representing the spread of different feature values. For categorical features $X^{cat}$, the distribution head is a multinomial distribution and each node outputs probabilities corresponding to different categories.

To train the autoencoder, we follow the common setting in Variational AutoEncoders~\citep{kingma2013auto}, minimizing 
as loss function the negative Evidence Lower-Bound (ELBO), defined as 
\begin{equation*}
    \mathcal{L}_{AE} =  \mathbb{E} _{\mathrm {z}   \sim q_{\mathcal{E} }(\mathrm {z}|\mathrm {x})}[-\log{p_{\mathcal{D} }(\mathrm {x}|\mathrm {z})]+D_{KL}(q_{\mathcal{E} }(\mathrm {z}|\mathrm {x})||p(\mathrm {z}))}.  
\end{equation*}
Here $q_{\mathcal{E} }(\mathrm {z}|\mathrm {x})$ is the posterior distribution of the latent space given the input $X$ after the encoder $\mathcal{E}$, and $p_{\mathcal{D} }(\mathrm {x}|\mathrm {z})$ is the output distribution of the decoder $\mathcal{D}$ given the latent space $Z$. $D_{KL}$ refers to the KL-divergence and $p(\mathrm {z})$ is a fixed prior distribution over the latent space $Z$. By setting $p(\mathrm {z})$ to a standard Gaussian distribution, the KL-divergence term serves as a regularizer that helps to avoid arbitrarily high-variance latent spaces.

\subsubsection{Latent Diffusion Model}

Once the input is mapped into the 
continuous latent space $Z$, a Gaussian diffusion process is the next 
component of the model. Within this process, for a latent variable $z^0$ generated by the encoder $\mathcal{E}$, the forward process in the diffusion model gradually adds Gaussian noise to the latent variable. Formally, with a total of $T$ timesteps and a predefined variance schedule $\beta^1,\dots,\beta^T$, the forward process at timestep $t$ is
\begin{equation*}
    q(z^t|z^{t-1})=\mathcal{N}(z^t;\sqrt{1-\beta ^t }z^{t-1}, \beta^t\mathbf{I}).  
\end{equation*}
Notably, the sampling $z^t$ at an arbitrary timestep $t$ can be expressed in closed form as
\begin{equation}
\label{eq:DMxt}
    q(z^t|z^0) = \mathcal{N}(z^t;\sqrt{\bar{\alpha }^t }z^0,(1-\bar{\alpha }^t)\mathbf {I}),  
\end{equation}
where $\alpha^t = 1 - \beta^t$ and $\bar{\alpha}^t =  {\textstyle \prod_{s=1}^{t} \alpha ^s} $.

The progressive forward process eventually converges to a pure noise space, characterized by a standard Gaussian distribution $p(z^T)=\mathcal{N}(z^T;\mathbf{0},\mathbf{I})$. Subsequently, the reverse process $p_\theta (z^{t-1} | z^t)$ is another Markov Chain with learned Gaussian transitions starting at $p(z^T)$. 

To learn the reverse denoising transitions, we adopt the methodology proposed in~\citep{ho2020denoising}. The crux of this approach involves estimating the added noise. Thus, the training objective of the diffusion component is formulated as minimizing the loss 
\begin{equation*}
    \mathcal{L}_{DF}=\mathbb{E} _{t,z^0,\epsilon}\left [   \left \| \epsilon - \epsilon _\theta (z^t,t  ) \right \|^2 \right ]. 
\end{equation*}
where $\epsilon$ is the true noise and $\epsilon_\theta$ is the estimated noise given the sampling $z^t$ and timestep $t$.

\subsection{Differential Privacy Framework}
\label{sec:DP}
To introduce a privacy protection in the latent space we employ the $f$-DP framework~\citep{dong2021gaussian} since it is capable to provide better bounds on the privacy leakage under composition, which is key in the training of neural models, which is done iteratively by means of stochastic gradient descent. These better bounds result in a more faithful privacy-utility tradeoff analysis. 

\subsubsection*{\textbf{$f$-DP background}}
In our paper, we adopt the $f$-DP framework to elevate privacy protection. This approach offers a clearer and more intuitive privacy explanation, encapsulating all necessary details to derive established DP metrics. Moreover, $f$-DP achieves a tighter privacy bound than traditional $(\varepsilon, \delta)$-DP, allowing for a more precise privacy evaluation~\citep{dong2021gaussian, wang2024unified}. 

Differential Privacy (DP), as introduced by Dwork et al.~\citep{dwork2006calibrating}, is a foundational framework for preserving the privacy of individuals' data within datasets. It quantifies the impact of an individual's data on the output of a randomized algorithm, ensuring minimal influence and thus protecting privacy.

In the $(\varepsilon, \delta)$-DP framework, 
a randomized mechanism $\mathcal{M}: \mathbb{D} \rightarrow \mathbb{R}$, where $\mathbb{D}$ is the domain and $\mathbb{R}$ the range, achieves $(\varepsilon, \delta)$-DP if for any two neighboring datasets $D$ and $D'$, differing by only a single record, it holds that 
\begin{equation*}
    \Pr(\mathcal{M}(D) \in S) \le e^{\varepsilon} \Pr(\mathcal{M}(D') \in S) + \delta,
\end{equation*}
where $S$ is a subset of possible outputs. The parameters $\varepsilon$ and $\delta$ quantify the privacy level, with lower values indicating stronger privacy guarantees.

Transitioning from traditional $(\varepsilon, \delta)$-DP analysis, the $f$-DP framework, proposed in~\citep{dong2021gaussian}, offers a refined perspective that relies on 
framing the adversary's challenge as a hypothesis testing problem. 
This framework introduces a trade-off function $f$ that represents the trade-off between false negatives (FN) and false positives (FP) in distinguishing between datasets $D$ and $D'$.

The FN and FP errors are defined as
\begin{equation*}
    \alpha_{\phi} = \mathbf{E}_{\mathcal{M}(D) \in S}[\phi(S)] \quad \text{and} \quad \beta_{\phi} = 1 - \mathbf{E}_{\mathcal{M}(D') \in S}[\phi(S)],
\end{equation*}
where $\phi \in [0,1]$ denotes the rejection rule applied to the output of the DP mechanism $\mathcal{M}$. The trade-off function is given by
\begin{equation*}
   \mathcal{T}(\mathcal{M}(D), \mathcal{M}(D'))(\alpha) = \inf_{\phi} \{ \beta_{\phi} : \alpha_{\phi} \leq \alpha \},
\end{equation*}
for a significance level $\alpha\in [0,1]$, 
signifying the optimal trade-off between FN and FP errors. 
A mechanism $\mathcal{M}$ is said to be $f$-DP if 
$\mathcal{T}(\mathcal{M}(D), \mathcal{M}(D')) \geq f$ for all 
neighboring datasets $D$ and $D'$.

Thanks to its functional definition, the $f$-DP framework can provide much tighter composition bounds than other existing definitions of DP. 
$f$-DP encompasses $(\epsilon,\delta)$-DP as a special case, wherein a mechanism is $(\epsilon,\delta)$-DP if and only if it conforms to $f_{\epsilon,\delta}$-DP, with $f_{\epsilon,\delta}(\alpha) =
\max \{ 0, 1-\delta - e^{\epsilon}\alpha, (1-\delta-\alpha)e^{-\epsilon}\}$. 

\subsubsection*{\textbf{DP-SGD}} 
 The Differentially Private Stochastic Gradient Descent algorithm (DP-SGD)~\citep{abadi2016deep} was designed for the differentially private training of neural networks. It achieves differential privacy by individually clipping (IC) the gradient of each individual sample within each mini-batch and adding Gaussian noise $\mathcal{N}(0, (C\sigma)^2\textbf{I})$ as to the gradient 
 \begin{equation}
     \tilde{g}_r \gets \frac{1}{|B|}  (  {\textstyle \sum_{i\in B}^{} [g_r(x_i)]_C 
     + \mathcal{N}(0, (C\sigma)^2\textbf{I})  } ).
     \label{eq:dpsgd}
 \end{equation}
 Here $[g_r(x_i)]_C=g_r(x_i)/\max_{}(1, \left \| g_r(x_i) \right \|_2 / C)$ is clipped from $g_r(x_i)$, the original gradient of sample $x_i$ at training round $r$, using the gradient norm bound $C$ and a mini batch size $B$. 

\subsubsection*{\textbf{Separation metric}}
To better illustrate the effectiveness of the $f$-DP guarantee, 
we introduce a novel metric called \textit{separation}, which intuitively indicates the strength of DP in the hypothesis testing trade-off by measuring the distance between the ideal and actual trade-off functions. 

Let $N$ be the dataset size, $b=\mathbb{E}[|B|]$ the sample (mini batch) size, and $\sigma$ the standard deviation of the Gaussian noise used in DP-SGD~\footnote{
We consider probabilistic sampling as in the Opacus library~\citep{yousefpour2021opacus} and use noise parameter $C\cdot \sigma$ and normalize with $C$ rather than $2C$.} as in~\eqref{eq:dpsgd}.   
Thus, $N/b$ equals the number of rounds in a single epoch and letting $E$ denote the total number of epochs, the total number of rounds is $R=(N/b)\cdot E$. 

Then DP-SGD is $C_{b/N}(G_{\sigma^{-1}})^{\otimes R}$-DP where $C_{b/N}$ is an operator representing the effect of subsampling in DP-SGD, $G_{\sigma^{-1}}$ is a Gaussian trade-off function characterizing the differential privacy (called Gaussian DP) due to adding Gaussian noise in DP-SGD, and the operator $\otimes R$ describes composition (of privacy leakage) over $R$ rounds. 


Following the asymptotic analysis in~\citep{dong2021gaussian}, 
DP-SGD converges to a $\mu$-Gaussian DP defined as 
\begin{equation*}
    G_{c\cdot h(\sigma)}\mbox{-DP \ \ for \ \ } c=\sqrt{bE/N},  
\end{equation*}
where the function $h(\sigma)$ is calculated as 
\begin{equation*}
h(\sigma) = \sqrt{2\left(e^{\sigma^{-2}}\Phi\left(\frac{3}{2\sigma}\right)+3\Phi\left(-\frac{1}{2\sigma}\right)-2\right)}. 
\end{equation*}

The ideal trade-off function is defined as $f(\alpha) = 1 - \alpha$, representing random guessing by the adversary; hence, it implies no privacy leakage. Since optimal trade-off functions are symmetric around the diagonal, separation between $1 - \alpha$ and $G_{\mu}(\alpha)$ can be measured as Euclidean distance between the point $(\frac{1}{2}, \frac{1}{2})$ on the curve $1 - \alpha$ and the point $(a, a)$ on the curve $G_{\mu}(\alpha)$, i.e., where $G_{\mu}(a) = a$. Here, $G_{\mu}(a) = \Phi(\Phi^{-1}(1 - a) - \mu)$, with $\Phi(\cdot)$ being the cumulative distribution function of the standard normal distribution. Thus the separation is denoted as 
\begin{equation}\label{eq:sep}
sep = \sqrt{2} \left| a - \frac{1}{2} \right|, \quad \text{s.t. } G_{\mu}(a) = a.
\end{equation}
For instance, taking the separation as 0.1, the $\mu$ calculated is 0.3563, and the distance between the trade-off function and the ideal curve is illustrated in Figure~\ref{fig:trade-off}.

\begin{figure}[h] 
     \centering
     \includegraphics[width=.4\linewidth]{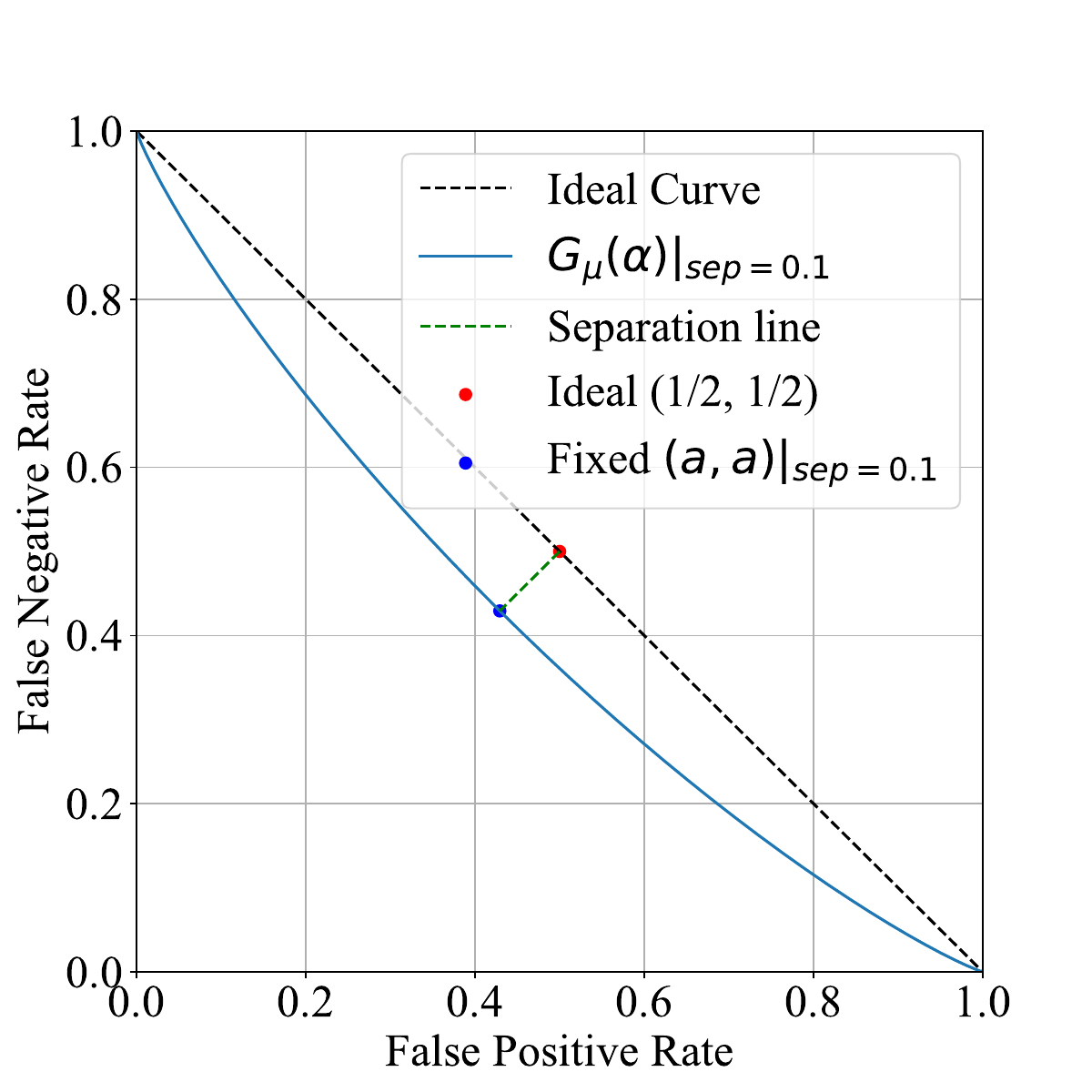}
    \caption{The separation between the ideal curve and the trade-off function.}
    \label{fig:trade-off}
 \end{figure}



We notice that DP guarantee is influenced by three hyperparameters: $\sigma$, $\frac{N}{b}$, and $E$. Clearly, given a target utility, smaller values of $E$ and larger values of $\frac{N}{b}$ enhance privacy protection. Based on these observations, we introduce the separation value as a novel term to evaluate privacy, which provides an intuitive explanation of the strength of DP. 




\subsection{Two-stage DP-SGD Training}
\label{sec:DPsgd}


Instead of using the traditional individual clipping (IC) as in~\eqref{eq:dpsgd}, as pointed out in \citep{nguyen2023batch}, 
a better way is to utilize batch clipping (BC) for DP training, i.e., 
\begin{equation}
    \tilde{g}_r \gets    {\textstyle \left[\frac{1}{|B|} \sum_{i\in B} g_r(x_i)\right]_C 
    + \mathcal{N}(0, (C\sigma)^2\textbf{I})  },
    \label{eq:batchdpsgd-2}
\end{equation}
In batch clipping, the average of the gradients within a batch is computed before applying clipping, as opposed to \eqref{eq:dpsgd}, which averages a sum of clipped individual gradients. This offers two key advantages.  
First, batch clipping allows for efficient computation of the sum of gradients across the entire mini-batch during both the forward and backward passes, thereby enhancing computational efficiency compared to individual clipping, which requires gradient computation for every single sample.  

\begin{algorithm}[htp]
\caption{DP enhanced two-stage training in \algo}
\label{algo:two-step}
\begin{algorithmic}[1]

\State \textbf{Input}: Tabular data $X = \{x_1, \ldots, x_N\}$, epochs $E_1$, $E_2$, batch sizes $B_1$, $B_2$, noise scale $\sigma$, norm bound $C$, timestep $T$
\State \textbf{Output}: Encoder $\mathcal{E}$, decoder $\mathcal{D}$, noise network $\epsilon_\theta$
\State \textbf{Initialize} $\mathcal{E}$, $\mathcal{D}$, $\epsilon_\theta$
\For{$e_1 = 1$ to $E_1$} \For{$r_1 = 1$ to $\lceil N / B_1 \rceil$} 
    \State Compute $\bar{g}_{r_{1_{\mathcal{E}}}} \gets \frac{1}{B_1} \nabla_\mathcal{E} \mathcal{L}_{AE}(\mathcal{E}, X_{r_1})$, 
    $\bar{g}_{r_{1_{\mathcal{D}}}} \gets \frac{1}{B_1} \nabla_\mathcal{D} \mathcal{L}_{AE}(\mathcal{D}, X_{r_1})$
\State Clip and add noise for $\mathcal{E}$: 
$\tilde{g}_{r_{1_{\mathcal{E}}}} \gets \frac{\bar{g}_{r_{1_{\mathcal{E}}}}}{\max(1, \|\bar{g}_{r_{1_{\mathcal{E}}}}\|_2 / C)} + \mathcal{N}(0, (C\sigma)^2\mathbf{I})$
\State Clip and add noise for $\mathcal{D}$: 
$\tilde{g}_{r_{1_{\mathcal{D}}}} \gets \frac{\bar{g}_{r_{1_{\mathcal{D}}}}}{\max(1, \|\bar{g}_{r_{1_{\mathcal{D}}}}\|_2 / C)} + \mathcal{N}(0, (C\sigma)^2\mathbf{I})$

\EndFor \EndFor
\State $Z^0 = \mathcal{E}(X)$
\For{$e_2 = 1$ to $E_2$} \For{$r_2 = 1$ to $\lceil N / B_2 \rceil$}
    \State Sample $Z_{r_2}^0 \sim q(Z^0)$, $t \sim \text{Uniform}(\{1, \ldots, T\})$, $\epsilon \sim \mathcal{N}(0, \mathbf{I})$
    \State Compute $g_{\epsilon_\theta} \gets \nabla_{\epsilon_\theta} \mathcal{L}_{DF}(\epsilon, \epsilon_\theta, t, Z_{r_2}^0)$ and update $\epsilon_\theta$
\EndFor \EndFor
\end{algorithmic}

\end{algorithm}

Second, batch clipping enables training Batch Normalization Layers in neural networks with a robust DP guarantee. As highlighted in~\citep{nguyen2023batch}, current implementations for IC that use batch normalization on the extensive training dataset lead to correlations among the updates across training rounds. Since these correlations are not considered, IC does not yield a solid DP guarantee from a theoretical perspective. However, batch normalization with BC over corresponding mini-batches can provide a solid DP argument within the $f-$DP framework. 

Therefore, based on the mentioned advantages, we employ DP-SGD with batch clipping to enhance the differentially private training of our latent tabular diffusion model, as presented in Algorithm~\ref{algo:two-step}. The training procedure consists of two steps. First, given a privacy budget, we train the Autoencoder component utilizing batch clipping alongside the injection of DP noise. Second, the DP-trained encoder generates the latent features of the original data, and the Gaussian diffusion model is trained on this latent feature space.

\section{Performance Evaluation}

In this section, we evaluate the proposed \algo on the aforementioned four datasets, employing the same quality and privacy risk metrics used in Section~\ref{sec:EmpiricalAnalysis}. We aim to answer if \algo can take advantage of the privacy protection from the DP mechanism without degrading the synthetic data quality. We specifically compare \algo against two other baselines, DP-CTGAN and DP-TabDDPM, by applying the DP-SGD training algorithm on CTGAN and TabDDPM, which represent the state-of-the-art GAN and diffusion-based generative models, respectively. 

\begin{table}[htb]
\centering
\resizebox{0.75\textwidth}{!}{%
\begin{tabular}{c|c|ccc|ccccc}
\toprule[0.9pt]
\multirow{2}{*}{Dataset} & \multirow{2}{*}{Method} & \multicolumn{3}{c|}{Quality Score $\uparrow$}   & \multicolumn{4}{c}{Privacy Risk $\downarrow$} \\ \cline{3-9} 
 &     & Resem. & Discri. & Utility  & S-out & Link & AIA   & MIA   \\ \bottomrule[0.9pt]
\multirow{3}{*}{Loan}        & TLDM       & 96          & 98               & 100     & 22.86        & 1.42        & 21.94 & 42.86 \\ \cline{2-9} 
& DP-CTGAN   & 40          & 11               & 54      & 0            & 0.15        & 1.18 & 2.86  \\ \cline{2-9} 
                            & DP-TabDDPM & 40          & 9                & 55      & 0            & 0.13        & 2.43 & 10.48 \\ \cline{2-9} 

                            & DP-TLDM    & 63          & 57               & 63      & 16.48        & 0.63        & 2.7  & 5.72  \\ \bottomrule[0.9pt]
\multirow{3}{*}{Housing}                                & TLDM       & 98          & 98               & 85      & 2.53         & 0.12        & 0.98  & 97.14 \\ \cline{2-9} 
& DP-CTGAN   & 37          & 9                & 21      & 0.34         & 0.01        & 0.05 & 8.58  \\ \cline{2-9} 
                            & DP-TabDDPM & 47          & 9                & 8       & 0.24         & 0.14        & 0.62 & 4.48  \\ \cline{2-9} 

                            & DP-TLDM    & 86          & 81               & 30      & 1.22         & 0.09        & 0.81 & 10.48 \\ \bottomrule[0.9pt]
\multirow{3}{*}{Adult}     
                            & TLDM       & 95          & 88               & 100     & 18.80        & 0.82        & 2.52  & 80.00 \\ \cline{2-9} 
                            & DP-CTGAN   & 44          & 10               & 49      & 28.36        & 0.17        & 0.75 & 8.58  \\ \cline{2-9} 
                            & DP-TabDDPM & 49          & 9                & 48      & 0.3          & 0.27        & 2.7  & 5.72  \\ \cline{2-9} 
                            & DP-TLDM    & 77          & 63               & 58      & 12.72        & 0.14        & 2.27 & 14.28 \\ \bottomrule[0.9pt]
\multirow{3}{*}{Cardio}     & TLDM       & 100         & 95               & 100     & 68.52        & 0.39        & 18.51 & 97.14 \\ \cline{2-9}
                            & DP-CTGAN   & 53          & 14               & 51      & 38.13        & 0.03        & 1.58 & 14.28 \\ \cline{2-9} 
                            & DP-TabDDPM & 43          & 9                & 71      & 0.99         & 0.15        & 1.06 & 0     \\ \cline{2-9}
                            & DP-TLDM    & 86          & 50               & 91      & 17.25        & 0.05        & 2.1  & 15.24 \\ \bottomrule[0.9pt]
\end{tabular}%
}
\caption{Impact of DP-SGD training on DP-CTGAN, DP-TabDDPM, and the proposed \algo.  Here, ``Resem." stands for Resemblance, ``Distrim." refers to Discriminability, ``S-out" denotes singling out attacks, and ``Link" represents linkability attacks.}
\label{tab:with_dp}
\end{table}

\textbf{Evaluation setup.} Identical to Section~\ref{sec:EmpiricalAnalysis}, we use resemblance, discriminability and utility to measure synthetic data quality. We evaluate the privacy risk, ranging from 1-100, for four types of attacks, singling out, linkability, AIA and MIA. The privacy measure is the theoretical separation value in~\eqref{eq:sep}, representing the maximum difference of typeI-typeII error between the random guess and DP-SGD, illustrated in Fig~\ref{fig:trade-off}. Three separation values are evaluated, namely $[0.1, 0.15, 0.2]$, where lower values indicate a stronger privacy level. We conduct the DP-SGP training on each generator under a given $\sigma$ value until the budget of separation depletes. For a fair comparison, we also apply batch clipping on all three synthesizers as outlined in~\eqref{eq:batchdpsgd-2}.

\begin{figure*}[htb]
    \centering
    \includegraphics[width=0.4\linewidth]{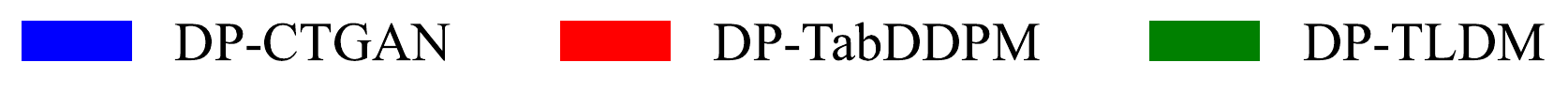} 
    \vspace{0.01cm} 
    
    \begin{subfigure}{0.49\textwidth}
        \includegraphics[width=\linewidth]{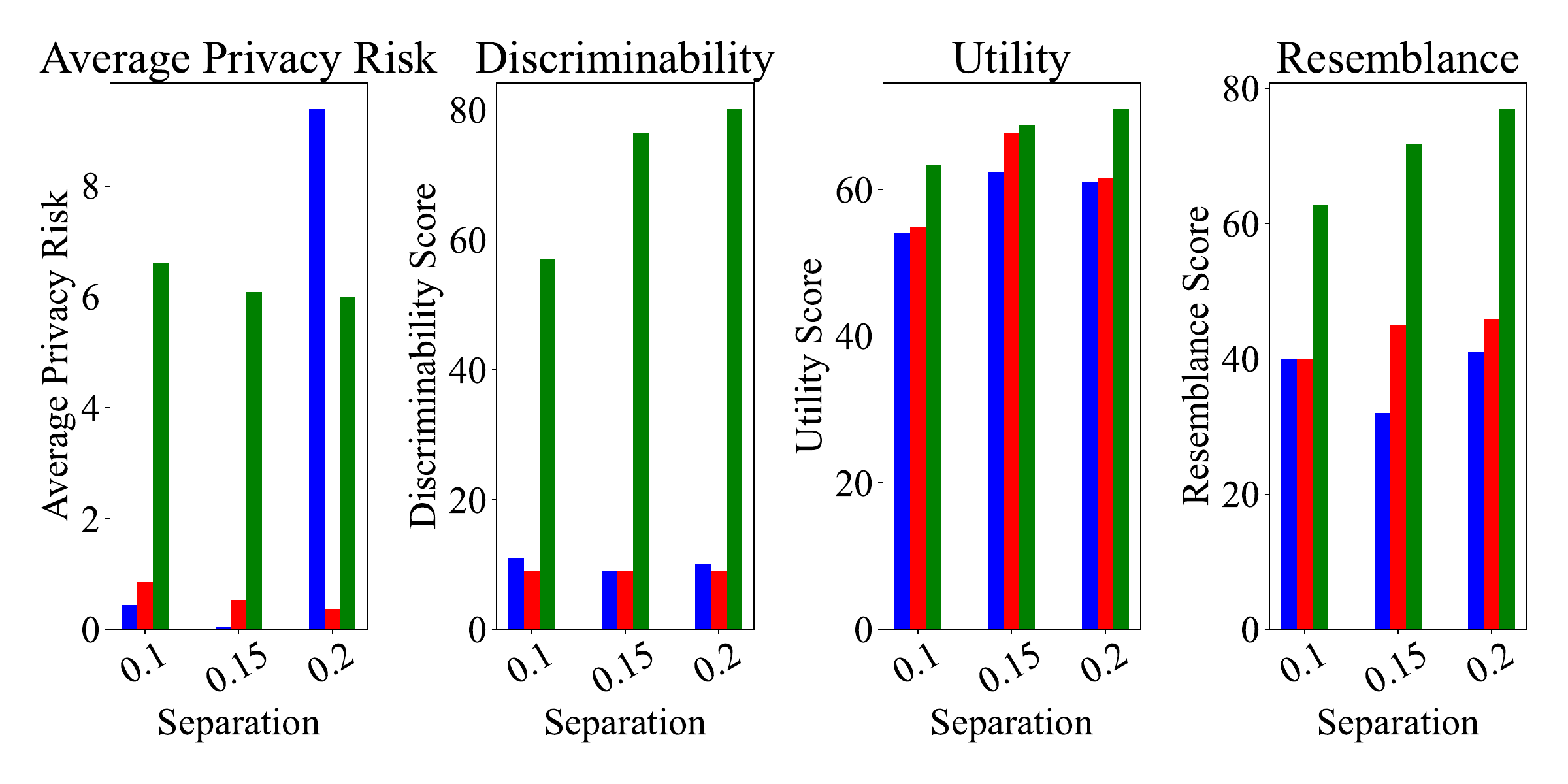}
        \caption{Loan}
        \label{fig:sub1}
    \end{subfigure}
    \hfill 
    \begin{subfigure}{0.49\textwidth}
        \includegraphics[width=\linewidth]{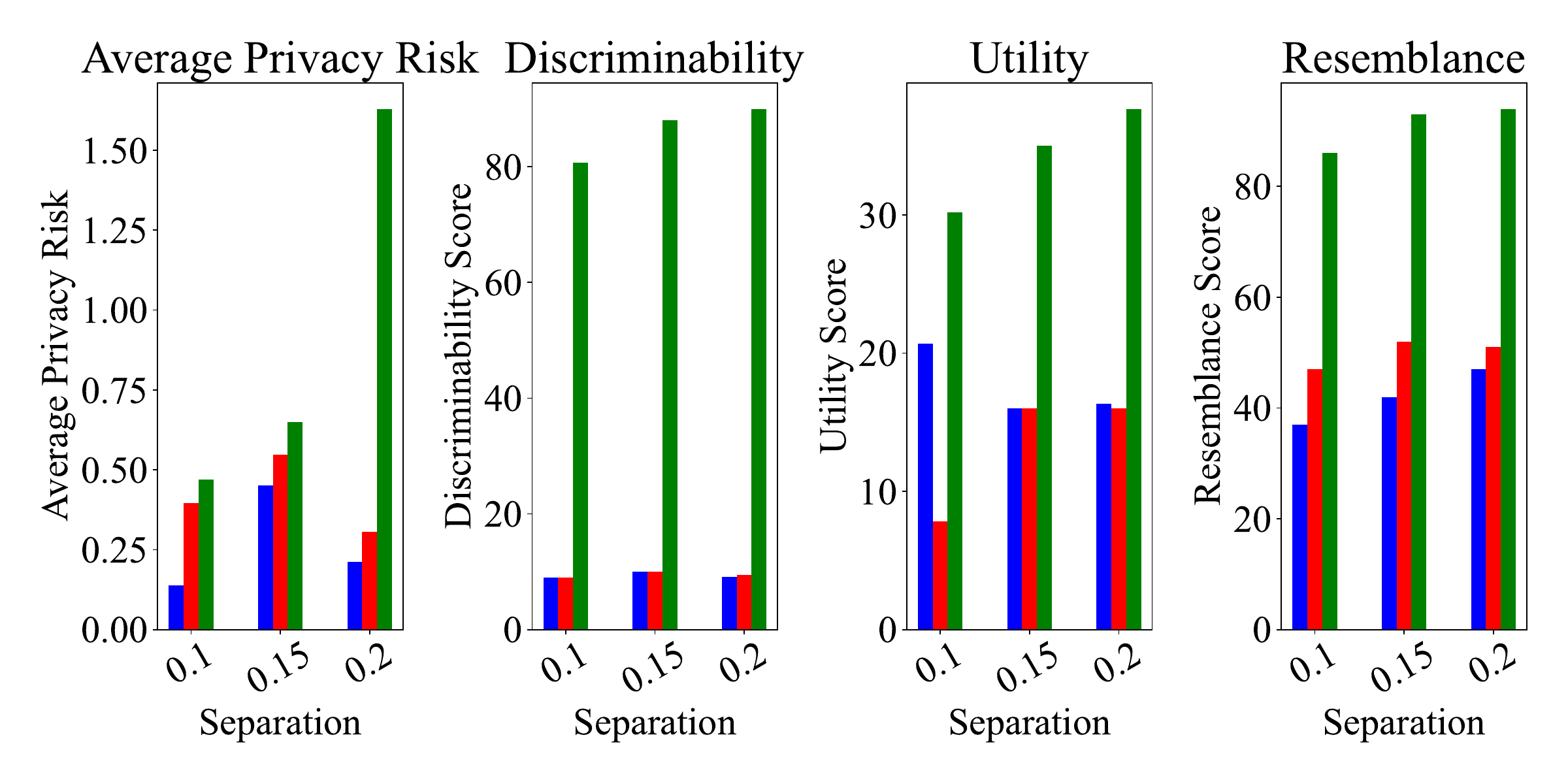}
        \caption{Housing}
        \label{fig:sub2}
    \end{subfigure}
    
    \begin{subfigure}{0.49\textwidth}
        \includegraphics[width=\linewidth]{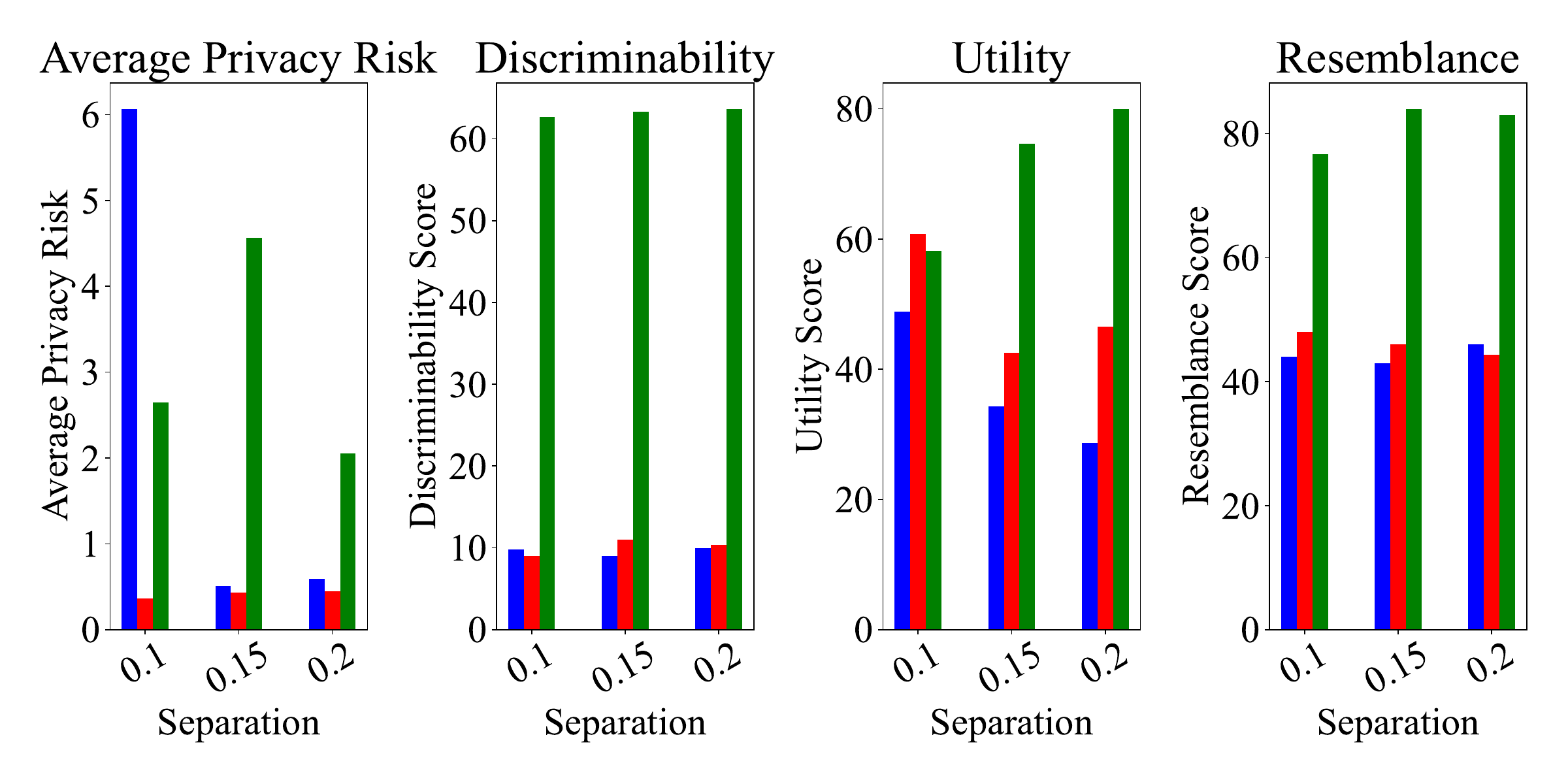}
        \caption{Adult}
        \label{fig:sub3}
    \end{subfigure}
    \hfill 
    \begin{subfigure}{0.49\textwidth}
        \includegraphics[width=\linewidth]{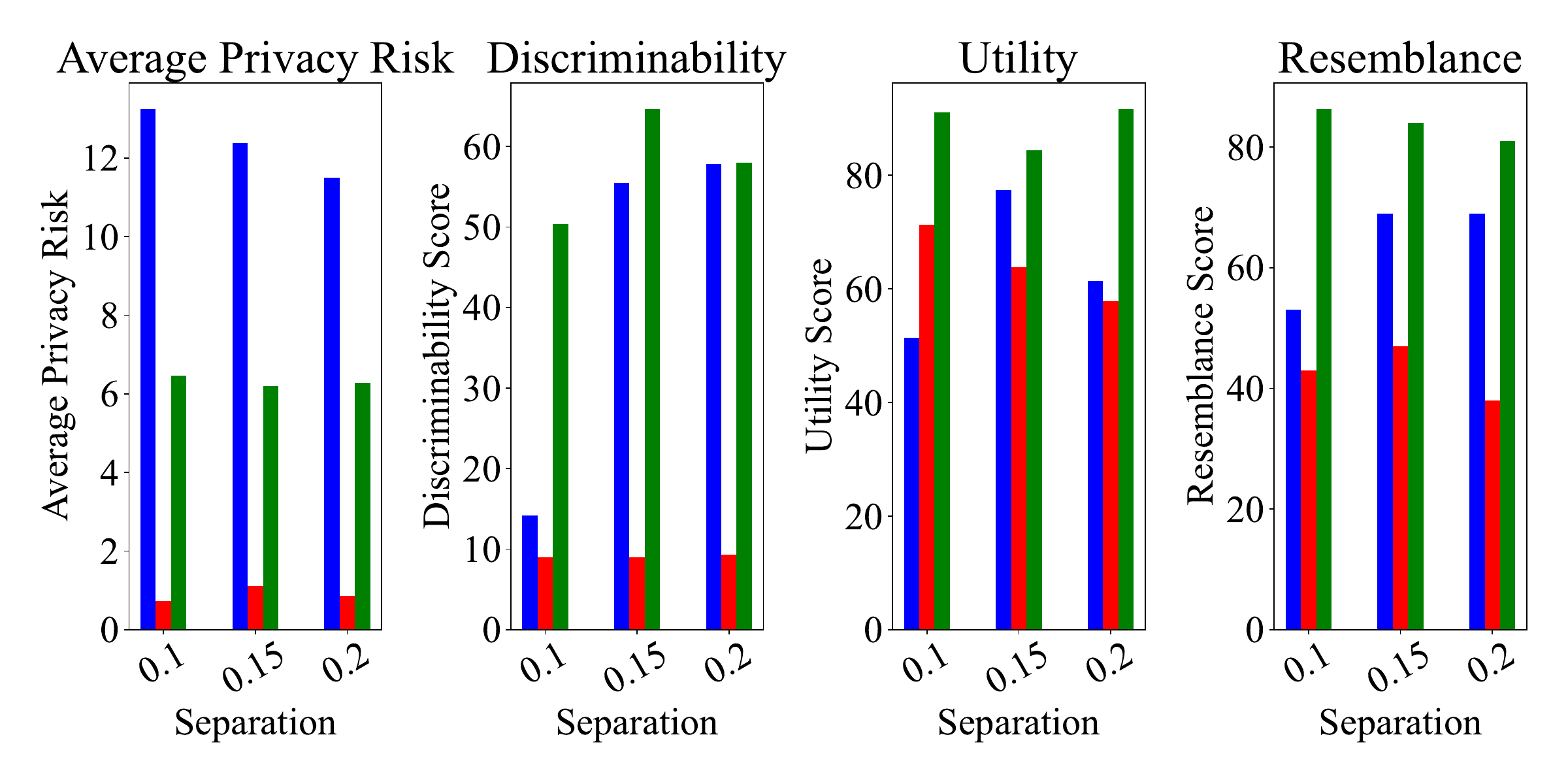}
        \caption{Cardio}
        \label{fig:sub4}
    \end{subfigure}
    
    \caption{Comparison between three DP-enhanced synthesizers on various datasets.}
    \label{fig:VaringSep}
\end{figure*}

\subsection{Overview}
We first present the overall performance for DP-CTGAN, DP-TabDDPM, and \algo in Table~\ref{tab:with_dp}. The specific separation value is $0.1$, which is the most meaningful DP protection level in our evaluation.  We summarize the key observations as follows. 

\emph{\algo achieves the optimal balance between data quality and privacy risk mitigation}. Across all DP-protected synthesizers, \algo consistently demonstrates the most favorable trade-off. It excels at achieving the highest resemblance, discriminability and utility scores, with comparable empirical risks. In sharp contrast, the two baseline methods fail to achieve any meaningful data quality scores with DP added, while \algo outperforms its counterparts by up to 3X across all four datasets.


\emph{DP protection yields a notable reduction in the risk of MIA on DP-TabDDPM and \algo.}  Notably, among all attacks considered, the most pronounced enhancement is found in MIA, where the risk diminishes substantially from approximately 90 to around 10. Given that MIA exploits additional information about the training dataset and model, its potential implications for the privacy of synthetic data are particularly severe. However, DP mechanism employed here effectively mitigates these risks, successfully defending against MIA.

\emph{A discernible reduction in privacy risks and data quality measures is evident when comparing DP and non-DP versions.} Across all three DP-protected synthesizers, the privacy risks demonstrably decrease at the expense of data quality, compared with their non-DP versions. This phenomenon is observed across all four datasets and against all four attacks. Particularly noteworthy is the significant enhancement observed in the cardio dataset. Specifically, notable improvements are observed in the Singling Out attack (risk decreases from an average of 60 to approximately 20), AIA (from an average of 20 to 2), and MIA (from an average of 90 to 10).

\emph{\algo exhibits the highest resilience to the DP mechanism considering data quality.}. Among all three data synthesizers, both DP-CTGAN and DP-TabDDPM experience substantial declines in data quality, particularly in discriminability, with scores dropping significantly from 92 (98) to 9 (9) in CTGAN (TabDDPM) on the housing dataset. In contrast, \algo manages to maintain a much higher data quality of synthetic data. We attribute the robust performance of \algo to its two-step training design. By implementing DP-SGD on the autoencoder networks and leveraging the diffusion backbone to offset the quality degradation in the autoencoder, \algo effectively preserves data utility despite the application of DP.

\subsection{Impact of privacy budget}

Here, we study the impact of varying separation values and summarize the results in Figure~\ref{fig:VaringSep}. The notion of Average Privacy Risk refers to the average risk score of Singling Out, Linkability and AIA. A higher separation value offers limited privacy protection but also introduces a lower perturbation to the quality of synthetic data. Consequently, we present the following noteworthy observations.

\emph{\algo consistently exhibits the best synthetic data quality across varying levels of privacy budget.} Across all four datasets, a distinct hierarchy emerges among the three synthesizers, with \algo surpassing DP-TabDDPM and DP-CTGAN. This can be explained by the two benefits of our two-stage training scheme: firstly, Diffusion Models (DDPM) inherently exhibit great resilience to noisy input~\citep{daras2024ambient, Hsieh2023}. By integrating the autoencoder with DP, the algorithm outputs latent representations with added perturbations. The resilience of the diffusion model ensures the generation of high-quality synthetic data. Secondly, the isolated two-stage training approach, where the privacy budget is solely allocated to the autoencoder stage, ensures that the diffusion process can refine and generate synthetic data without further compromising privacy. This efficient use of the privacy budget allows for the production of synthetic data that is not only of high quality but also adheres strictly to the required privacy constraints. The diffusion stage, not requiring additional privacy budget, acts as a compensatory mechanism for any potential decrease in data utility due to the privacy-preserving perturbations introduced in the autoencoder stage.



\emph{Across different datasets and separation values, \algo and DP-CTGAN generally have higher privacy risks}. However, the significantly superior data quality produced by \algo does result in greater privacy leakage. Nonetheless, considering that the privacy risk is quantified on a scale from 0 to 100, all datasets demonstrate that our model maintains a privacy risk below 8. This indicates that \textit{algo} successfully achieves an optimal balance between data quality and privacy protection.


Overall, these findings underscore that:
The two-stage training scheme of \algo, which leverages the inherent robustness of diffusion models to noisy inputs, achieves the optimal privacy-utility tradeoff among three DP-generators at equivalent privacy levels.

\section{Conclusion}
Motivated by the increasing adoption of synthetic tables as a privacy-preserving data sharing solution,
we design \algo, a latent tabular diffusion trained by DP-SGD, following the $f$-DP framework. Key components of \algo are i) an autoencoder network to transform  tabular data into a compact and unified latent representation, and ii) a latent diffusion model to synthesize  latent tables. 
Thanks to the two-component design, 
and by applying DP-SGD to train the autoencoder,  
\algo obtains a rigorous DP guarantee, measured by the separation value.    
Our evaluation results 
against tabular GAN and regular tabular diffusion models trained with DP-SGD show that \algo can effectively mitigate the empirical privacy risks of synthetic data while achieving  15-50\% higher data quality than other synthesizers with a stringent theoretical privacy budget.

\section*{Acknowledgments}
This research is part of the Priv-GSyn project, 200021E\_229204 of Swiss National Science Foundation, and the DEPMAT project, P20-22 / N21022, of the research programme Perspectief which is partly financed by the Dutch Research Council (NWO).

\bibliographystyle{splncs04}
\bibliography{sample_base}

\newpage

\appendix

\section{Nomenclature}
\begin{description}[leftmargin=1.0in, labelwidth=0.8in, labelsep=0.2in]
\item[$\alpha_{\phi}$, $\beta_{\phi}$] False negative and false positive rates, respectively, under a specific rejection rule $\phi$.
\item[$\bar{\alpha}^t$] Parameters defining variance and transformation across timesteps.
\item[$\alpha^t$] Parameters defining variance and transformation across timesteps.
\item[$b$] Expected sample (mini-batch) size in DP-SGD.
\item[$B$] Batch size in the training process.
\item[$\beta^t$] Parameters defining variance and transformation across timesteps.
\item[$C$] Gradient norm bound for clipping gradients during DP-SGD training.
\item[$D$, $D'$] Neighboring datasets that differ by only a single record.
\item[$\mathcal{D}$] Decoder component of the autoencoder, converting latent representation back to original data space.
\item[$\delta$] Probability parameter in $(\varepsilon, \delta)$-DP, allowing for the privacy guarantee to be violated with a small probability.
\item[$\mathbb{D}$] Domain of the randomized mechanism, representing the dataset space.
\item[$E$] Total number of training epochs in DP-SGD.
\item[$\mathcal{E}$] Encoder component of the autoencoder, transforming input data into a continuous latent representation.
\item[$E_1, E_2$] Training epochs for the autoencoder component and the latent diffusion component, respectively.
\item[$\epsilon$, $\epsilon_\theta$] True and estimated noise in the diffusion process.
\item[$\varepsilon$] Privacy loss parameter in $(\varepsilon, \delta)$-DP, controlling the allowable increase in output likelihood due to a change in a single record.
\item[$f$] Trade-off function in the $f$-DP framework, representing the balance between false negatives and false positives in distinguishing between datasets.
\item[$G_{\sigma^{-1}}$] Gaussian trade-off function characterizing differential privacy due to adding Gaussian noise in DP-SGD.
\item[$\tilde{g}_r$] Noisy gradient after applying batch clipping and adding Gaussian noise during DP-SGD training.
\item[$\mathcal{L}_{AE}$, $\mathcal{L}_{DF}$] Loss functions for autoencoder and diffusion components, respectively.
\item[$\mathcal{M}$] Randomized mechanism used in differential privacy.
\item[$\mu$] Parameter defining the strength of the Gaussian DP guarantee in DP-SGD.
\item[$\mu$-Gaussian DP] A measure of differential privacy based on Gaussian differential privacy, parameterized by $\mu$.
\item[$N$] Dataset size.
\item[$\mathcal{N}(0, (C\sigma)^2\mathbf{I})$] Gaussian noise distribution with mean 0 and variance scaled by $(C\sigma)^2$.
\item[$p(x_T)$] Distribution at the final step $T$, aiming for a simple form like Gaussian.
\item[$p_{con}(x_T)$, $p_{dis}(x_T)$] Distributions for continuous and categorical features at step $T$, respectively.
\item[$p_\theta (x_{t-1} | x_t)$] Reverse process, estimating data recovery from noise.
\item[$q(x_t | x_{t-1})$] Forward process at step $t$, modeling transition probabilities.
\item[$R$] Total number of training rounds in DP-SGD, calculated as $(N/b) \cdot E$.
\item[$\mathbb{R}$] Range of the randomized mechanism, representing the output space.
\item[$sep$] Separation metric measuring the distance between the ideal trade-off function and the actual trade-off function in the $f$-DP framework.
\item[$\sigma$] Noise scale used for adding Gaussian noise in the DP-SGD algorithm to ensure differential privacy.
\item[$T$] Total number of steps in the diffusion process.
\item[$\mathcal{T}(\mathcal{M}(D), \mathcal{M}(D'))(\alpha)$] Trade-off function capturing the optimal balance between false negatives and false positives for distinguishing between $D$ and $D'$.
\item[$X$] Original tabular data containing both continuous and categorical features.
\item[$X^{cat}$] Categorical features within the tabular data.
\item[$X^{con}$] Continuous features within the tabular data.
\item[$\tilde{X}$] Reconstructed data from the latent representation by the decoder.
\item[$x_0$] Original data before the diffusion process.
\item[$Z$] Continuous latent representation of original data by encoder.
\item[$z^0$, $z^t$] Latent variables at initial and timestep $t$, within the diffusion model.
\end{description}
\section{Risk-Utility Quantification}
\label{sec:privacyRisk}

In this section, we introduce our risk-utility quantification framework, as illustrated in Figure~\ref{fig:framework}. Given an original dataset, the synthesizers generate synthetic data, which is then assessed from two crucial perspectives: utility and privacy risk.

For utility quantification, three metrics will be reported: i) resemblance, 
ii) discriminability, 
and iii) utility. 
Below we provide further details on the on the synthesizers employed and the utility metrics. 
Regarding privacy risk, we consider four distinct attacks: i) singling out, ii) linkability, iii) attribute inference attack (AIA), and iv) membership inference attack (MIA), to measure different dimensions of privacy risks in the synthetic data. 
Below we describe these attacks and the associated metrics. 

\begin{figure*}[htbp]
    \centering
    \includegraphics[width=0.99\textwidth]{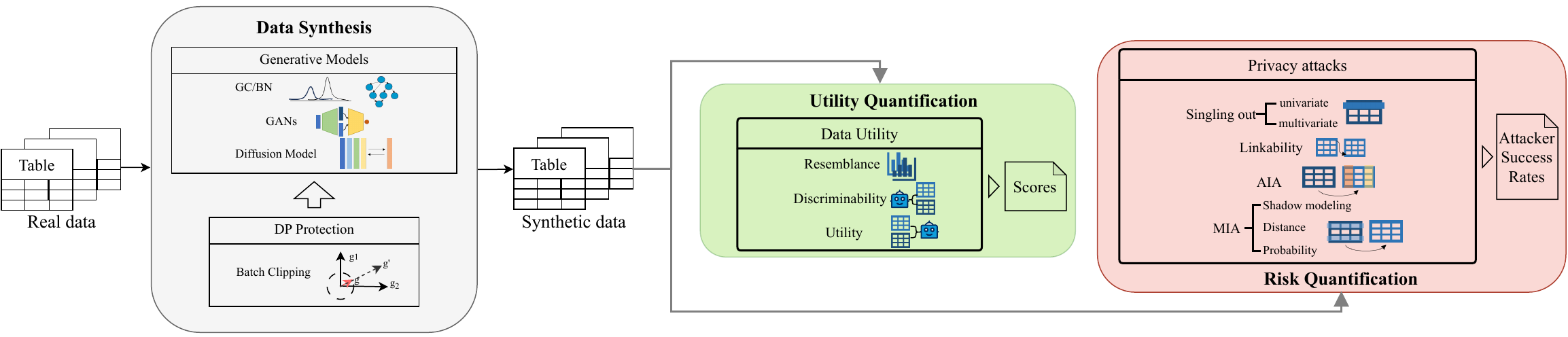}
        \caption{The risk-utility quantification and enhancing framework: from training to generation. Starting with the original tabular data, synthetic data is generated by synthesizers with or without DP protection. This synthetic data is then evaluated from two critical perspectives: utility (resemblance, discriminability, and utility) and privacy risk (singling out, linkability, attribute inference attack (AIA), and membership inference attack (MIA)). \lc{do we keep the DP here? Can we further improve this figure? }}
    \label{fig:framework}
\end{figure*}

\subsection{Generative models}
\label{sec:gen_models}
We employ six generative models in our framework, including GAN-based, statistical and diffusion-based models, described next. 


\subsubsection*{GAN-based models} 
We consider three different Generative Adversarial Networks (GAN) models. 
First, \textbf{CTGAN}~\citep{xu2019modeling} employs GANs with a focus on conditional generation. It employs mode-specific normalization to handle non-Gaussian 
distributions in continuous columns, and a conditional generator 
to address class imbalance in categorical columns. 
Second, \textbf{CopulaGAN}~\citep{sdvdoc} improves upon CTGAN by utilizing cumulative distribution function-based transformations with Gaussian Copulas. 
It also performs inference using a likelihood approach, enhancing CTGAN's ability to learn real data trends. 
Third, \textbf{ADS-GAN}~\citep{yoon2020anonymization} is a conditional GAN framework that generates synthetic data while minimizing re-identification risk. It achieves a certain degree of anonymization by incorporating a record-level identifiability metric 
into the generator's loss function. 
\subsubsection*{Statistical models} 
Here we consider Gaussian Copula models. 
In the \textbf{Gaussian Copula (GC)} method~\citep{nelsen2006introduction}, the training data is used to obtain a 
Gaussian joint probability distribution that captures both marginal distributions and interdependence structures. 

\subsubsection*{Difussion models} 
Diffusion models have recently become the leading paradigm in generative models for computer vision and NLP. 
In our framework, we consider~\textbf{TabDDPM}~\citep{kotelnikov2023tabddpm}, 
which extends diffusion models to tabular datasets, outperforming existing GAN/VAE alternatives. 
It employs the Gaussian diffusion process, a key component of the original DDPM~\citep{ho2020denoising}, to model numerical columns effectively. 
It also uses the multinomial diffusion process to model categorical and binary features and to introduce uniform noise across classes to corrupt data. 

The above-mentioned models are implemented following two Python libraries for tabular data synthesis: i) we employ Synthetic Data Vault~\footnote{https://github.com/sdv-dev/SDV} for CTGAN, CopulaGAN and Gaussian Copula, and ii) Synthcity~\footnote{https://github.com/vanderschaarlab/synthcity} for Bayesian Network, ADS-GAN and TabDDPM. To ensure a fair comparative analysis, neural networks used across all models have the same architecture consisting of three multi-layer perceptron (MLP) layers, each comprising 256 dimensions.

\subsection{Utility metrics}
\label{sec:utility_metrics}
To evaluate the quality of the synthetic data, we use three metrics, namely resemblance, discriminability, and utility, to assess whether the synthetic results are similar to the original data as well as practically useful. These metrics follow 
common practice in synthetic data generation, 
and are reported as scores in the 0-100 range.  

\subsubsection*{\textbf{Resemblance}}
The resemblance metric measures how closely the distribution and inter-correlation of the columns in the synthetic data match the original data, ensuring that the synthetic data captures the statistical patterns and characteristics of the original data. 
Our resemblance metric is composed of five similarity measures: 
\begin{itemize}[leftmargin=0.2cm]
\item \textbf{Column Similarity} calculates the correlation between each original and synthetic column, using Pearson's coefficient for numerical columns and Theil's U for categorical columns. 
\item \textbf{Correlation Similarity} measures the correlation between the correlation coefficients of each column pair. 
First, the Pearson correlation for numerical pairs, Theil's U for categorical pairs, and the correlation ratio for mixed 
cases are calculated. Then, the correlation between these coefficients 
is calculated.
\item \textbf{Statistical Similarity} employs Spearman's Rho to correlate descriptive statistics (minimum, maximum, median, mean, and standard deviation) of numerical columns in synthetic and original data.  
\item \textbf{Jensen-Shannon Similarity} uses the Jensen-Shannon distance 
between the probability distributions of the original and synthetic columns. 
One minus this distance is used so that higher scores are better, as in the other metrics. 
\item \textbf{Kolmogorov-Smirnov Similarity} uses the Kolmogorov-Smirnov distance to measure the maximum difference between the cumulative distributions of each original and synthetic column. 
Once again, one minus the distance is used so that a higher score is better.
\end{itemize}

\subsubsection*{\textbf{Discriminability}}
This metric measures how closely the synthetic data resembles the real data such that a binary classifier (XGBoost) cannot differentiate between the two.
We measure this with the mean-absolute error between the classifier's probabilities and the uniform distribution (50\% probability for either class), which is 0 when the classifier cannot distinguish between the two datasets.  
One minus the mean-absolute error is used so that higher scores are better. 

\subsubsection*{\textbf{Utility}}
Utility measures how well the synthetic data performs like the original data in downstream machine learning tasks. 
For each column, a classifier or regressor (XGBoost) is trained with 3-fold cross-validation to predict the column from the remaining columns. Models are trained either on real or synthetic data, but, in both cases, evaluated on a hold-out set of real data. The downstream performance is calculated by taking the 90th percentile of macro-averaged F1 scores for categorical columns and D2 absolute error scores (clipped to 0 and 1) for continuous columns. The utility score is derived from the ratio of the downstream performance of the synthetic data to that of the real data.


\subsection{Threat model}
\label{sec:ThreatModel}
In our threat model, we elucidate the prior knowledge that an attacker needs to know for potential attacks, focusing on the synthesizer, synthetic data, and auxiliary data.

For synthesizer knowledge, the attacker is assumed to possess no information about the underlying mechanisms of the synthesizer, adhering to the realistic black-box scenario. Besides, for a conservative privacy risk assessment, we presume the attacker has full access to synthetic data, anticipating worst-case scenarios like public release or online API accessibility. These assumptions are deliberately chosen for a comprehensive and resilient privacy risk assessment, considering potential vulnerabilities in worst-case situations. 
Concerning auxiliary data, the necessary information varies for different attacks. For singling out, no prior knowledge of auxiliary data is required. However, for linkability, attribute inference, and membership inference attacks, the attacker is assumed to know the target records $T$, randomly drawn from the training data. 

In particular, for linkability attacks, the needed auxiliary data comprises two disjoint sets of attributes $T[:, A]$ and $T[:, B]$, extracted from $T$. For attribute inference attacks, the attacker needs to know the values of a set of attributes $T[:, A]$. 
In shadow modeling-based membership inference attacks, the attacker, alongside the target set $T$, has access to another reference dataset $X_R$. 
As in~\citep{stadler2022synthetic}, this dataset mirrors the distribution of the training dataset and may or may not have overlapping records with it.

\subsection{Attacks}
As mentioned before, to quantify the privacy risks of tabular data synthesizers, we employ four distinct attacks in our evaluation framework. Three of these attacks—\textbf{singling out}, \textbf{linkability}, and \textbf{attribute inference attack (AIA)}—are derived from the guidelines set by the European General Data Protection Regulation (GDPR), following~\citep{giomi2023unified}. 
Additionally, we incorporate the \textbf{membership inference attack (MIA)}, a well-established but previously omitted facet in~\citep{giomi2023unified}. This enhances our privacy quantification framework by allowing for the measurement of an additional and widely acknowledged dimension of privacy in synthetic data. 

The implementation of singling out, linkability and AIA follows the open-source library Anonymeter~\footnote{https://github.com/statice/anonymeter} from~\citep{giomi2023unified} while the membership inference attack (MIA) follows the TAPAS toolbox~\footnote{https://github.com/alan-turing-institute/tapas} from~\citep{houssiau2022tapas}. 

\subsubsection*{\textbf{Singling out}}
The singling out attack~\citep{giomi2023unified} aims to create predicates from the synthetic dataset that could identify an individual present in the training dataset. For example, if an attacker can determine that a dataset has only one individual with attributes like age: 25, height: 168, weight: 62, and cholesterol: 1, that individual is considered "singled out". 

Following the implementation in \citep{giomi2023unified}, two algorithms are applied, the \textbf{univariate algorithm} and the \textbf{multivariate algorithm}. Both algorithms are based on the intuition that unique values or combinations of unique values in the synthetic data may also be unique in the original data. 
In the univariate algorithm, unique values are sampled for each attribute to obtain a random selection of predicates. In the multivariate algorithm, this is done for full records to obtain multivariate predicates. 


\subsubsection*{\textbf{Linkability}}
The linkability attack~\citep{giomi2023unified} aims to associate two or more records. 
The linkability arises when an attacker has two disjoint sets of attributes $X_A$ (e.g., age and height) and $X_B$ (e.g., weight and cholesterol level) from the original dataset, 
so that it can use the synthetic dataset to determine that two records $x_a \in X_A$ and $x_b \in X_B$ belong to the same individual. 

The attack works with two disjoint sets of attributes $T[:, A]$ and $T[:, B]$ of the target set $T$, which it uses to identify the $k$ nearest neighbors for every record in $T[:, A]$ and $T[:, B]$. 
A link between $x_a$ and $x_b$ is established if they share at least one common neighbor.

\subsubsection*{\textbf{Attribute inference attack (AIA)}}
The AIA attack~\citep{giomi2023unified} involves deducing undisclosed attribute values from information in the synthetic dataset. If an attacker knows certain attributes of an individual, such as age: 25, height: 168, and weight: 62, they might use the synthetic dataset to infer the cholesterol level of the individual. 

Given $N$ target records characterized by a set of known attributes $T[:, A]$, the nearest neighbor algorithm is applied again to perform AIA attacks. For each target record, the attacker seeks the closest synthetic record within the subspace defined by the attributes in the target records. The values assigned to the secret attributes of this closest synthetic record serves as the attacker's guess.

\subsubsection*{\textbf{Membership inference attack (MIA)}}
The MIA attack~\citep{shokri2017membership} aims to determine if a specific data record is present in the training dataset. MIA attacks have gained substantial attention within the research community, leading to various proposed strategies for inferring the membership status of synthetic data points. 
In our framework, we employ three types of MIA strategies, based on shadow modeling, distance and probability. 
This diverse set accommodates a range of adversarial scenarios, recognizing different adversary capabilities and constraints.

In the \textbf{shadow modeling} approach, 
given a reference dataset $X_R$, which shares the same distribution as the training dataset, and a target record $x_t$, multiple training sets $X_i$ from $X_R$ are sampled and shadow models trained 
to generate synthetic datasets $X_s$ from $X_i$ and $X_s'$ from $X_i' = X_i \cup x_t$. 
A classifier 
is then trained on labeled synthetic datasets $X_s$ and $X_s'$ to predict the presence of target records in the training data $X_{train}$. 
To reduce the effect of high-dimensionality and sampling uncertainty, instead of directly training on $X_s$ and $X_s'$, the \textbf{NaiveGroundhog} strategy uses a basic feature set $F_{naive}$ for training, while \textbf{HistGroundhog} utilizes a histogram feature set $F_{hist}$ with marginal frequency counts of each data attribute~\citep{stadler2022synthetic}.

\textbf{Distance-based} MIA strategies, such as \textbf{Closest Distance-Hamming} and \textbf{Closest Distance-L2}, focus on identifying the local neighborhood of the target record within the synthetic dataset. The attacker predicts membership based on the distance between the target record and its nearest neighbor in the synthetic dataset, with an empirically selected threshold.

Finally, the \textbf{probability-based} \textbf{Kernel Estimator} 
uses a density estimator 
to fit synthetic data, employing the estimated likelihood to predict membership. If the likelihood surpasses a threshold, 
the target record is predicted to be a member of the training set.

In assessing MIA risks for $N$ target records, all five strategies are executed, and results associated with the highest privacy risk are reported to provide conservative risk analysis, accounting for the worst-case scenario.

\subsubsection*{\textbf{Metrics}} 
\label{sec:PrivacyMetrics}


\textbf{Relative Risk Indicator}
The attacker success rate is a common metric for Membership Inference Attacks (MIA). However, for singling out, likability, and attribute inference attacks, as noted by \citep{giomi2023unified}, there's a distinction: certain information might be inferred from patterns inherent in the entire population $X_{ori}$ rather than solely from the training dataset $X_{train}$ and its synthetic counterpart.

Following \citep{giomi2023unified}, the original dataset $X_{ori}$ is split into two disjoint partitions: $X_{train}$ and $X_{control}$. Privacy risk is then quantified by comparing attacker success rates for targets drawn from $X_{train}$ and $X_{control}$:
\begin{equation}
    R = \frac{\hat{\tau}_{train}-\hat{\tau}_{control}}{1-\hat{\tau}_{control}}
\end{equation}
Here, $\hat{\tau}_{train}$ represents the attacker success rate when targets are solely from $X_{train}$, and $\hat{\tau}_{control}$ when targets are from $X_{control}$.

The numerator in the equation compares the attack's efficacy against $X_{train}$ versus $X_{control}$. As there is no overlap between $X_{control}$ and $X_{train}$, successful inferences against it likely stem from population-wide patterns. By removing the contribution against $X_{control}$, we isolate the attack's performance against $X_{train}$ and its synthetic data. The denominator normalizes the ratio, depicting the maximum improvement over the control attack achievable by a perfect attacker ($\tau=1$). To standardize our MIA metric with other attacks, we establish a baseline assumption that $\tau_{control}$ for the MIA attack is 50\%. This assumption enables us to utilize privacy risk as a means to evaluate the MIA attack effectively.




\begin{table}[htb]
\centering
\begin{tabular}{cccc}
\toprule[0.9pt]
        & No. Rows & No. Columns & Missing Values \\ \bottomrule[0.9pt]
Loan    & 5000     & 14          & 0          \\ \hline
Housing & 20640     & 10           & 0.1\%          \\ \hline
Adult   & 48842    & 14          & 0.94\%           \\ \hline
Cardio  & 70000    & 13          & 0               \\  \bottomrule[0.9pt]
\end{tabular}
\caption{Characteristics of four tabular datasets used in our study.}
\label{tab:DataStatistics}
\end{table}

\section{Additional Results}
\subsection{Additional MIA Results}~\label{app:mia}
In the main section, we identify membership inference attacks (MIAs) as presenting the largest privacy risk, and designate them as the final MIA privacy risk assessment. Table~\ref{tab:mia_add} displays the privacy risks associated with all the attacks we have analyzed.

\begin{table}[htb]
\centering
\resizebox{0.8\textwidth}{!}{%
\begin{tabular}{c|c|c|ccccc}
\toprule[0.9pt]
\multicolumn{1}{l}{Dataset} & Method     & MIA   & NG    & HG    & CD-H  & CD-L  & KE    \\\bottomrule[0.9pt]
\multirow{9}{*}{Loan}       & CopulaGAN  & 2.86  & 0     & 0     & 0     & 2.86  & 2.86  \\ \cline{2-8} 
                            & CTGAN      & 2.86  & 0     & 0     & 2.86  & 0     & 0     \\\cline{2-8} 
                            & ADS-GAN    & 22.86 & 14.28 & 22.86 & 8.58  & 5.72  & 5.72  \\\cline{2-8} 
                            & GC         & 5.72  & 0     & 0     & 5.72  & 0     & 0     \\\cline{2-8} 
                            & TabDDPM    & 45.72 & 11.42 & 45.72 & 8.58  & 8.58  & 8.58  \\\cline{2-8} 
                            & TLDM       & 42.86 & 0     & 42.86 & 0     & 0     & 0     \\\cline{2-8} 
                            & DP-CTGAN   & 2.86  & 0     & 0     & 0     & 2.86  & 0     \\\cline{2-8} 
                            & DP-TabDDPM & 10.48 & 0     & 10.48 & 0     & 2.86  & 0     \\\cline{2-8} 
                            & DP-TLDM    & 5.72  & 0     & 5.72  & 0     & 0     & 2.86  \\\bottomrule[0.9pt]
\multirow{9}{*}{Housing}    & CopulaGAN  & 20    & 20    & 8.58  & 2.86  & 2.86  & 2.86  \\\cline{2-8} 
                            & CTGAN      & 20    & 0     & 0     & 20    & 0     & 0     \\\cline{2-8} 
                            & ADS-GAN    & 48.58 & 0     & 28.58 & 48.58 & 5.72  & 5.72  \\\cline{2-8} 
                            & GC         & 5.72  & 5.72  & 0     & 0     & 2.86  & 2.86  \\\cline{2-8} 
                            & TabDDPM    & 88.58 & 0     & 85.72 & 88.58 & 0     & 0     \\\cline{2-8} 
                            & TLDM       & 97.14 & 0     & 97.14 & 94.28 & 8.58  & 8.58  \\\cline{2-8} 
                            & DP-CTGAN   & 8.58  & 0     & 8.58  & 0     & 0     & 0     \\\cline{2-8} 
                            & DP-TabDDPM & 4.48  & 7.14  & 7.14  & 0     & 0     & 1.42  \\\cline{2-8} 
                            & DP-TLDM    & 10.48 & 0     & 10.48 & 0     & 2.86  & 0     \\\bottomrule[0.9pt]
\multirow{9}{*}{Adult}      & CopulaGAN  & 17.14 & 0     & 2.86  & 17.14 & 0     & 0     \\\cline{2-8} 
                            & CTGAN      & 10    & 0     & 0     & 10    & 0     & 0     \\\cline{2-8} 
                            & ADS-GAN    & 20    & 20    & 20    & 8.58  & 0     & 0     \\\cline{2-8} 
                            & GC         & 8.58  & 8.58  & 2.86  & 0     & 2.86  & 2.86  \\\cline{2-8} 
                            & TabDDPM    & 94.28 & 8.58  & 94.28 & 48.58 & 0     & 0     \\\cline{2-8} 
                            & TLDM       & 80    & 0     & 80    & 8.58  & 5.72  & 8.58  \\\cline{2-8} 
                            & DP-CTGAN   & 8.58  & 0     & 0     & 0     & 9.52  & 14.28 \\\cline{2-8} 
                            & DP-TabDDPM & 5.72  & 0     & 0     & 0     & 1.42  & 5.72  \\\cline{2-8} 
                            & DP-TLDM    & 14.28 & 8.58  & 0     & 5.72  & 15.24 & 14.28 \\\bottomrule[0.9pt]
\multirow{9}{*}{Cardio}     & CopulaGAN  & 0     & 0     & 0     & 0     & 0     & 0     \\\cline{2-8} 
                            & CTGAN      & 11.42 & 0     & 11.42 & 0     & 5.72  & 2.86  \\\cline{2-8} 
                            & ADS-GAN    & 31.42 & 2.86  & 31.42 & 2.86  & 0     & 0     \\\cline{2-8} 
                            & GC         & 22.86 & 22.86 & 0     & 0     & 0     & 0     \\\cline{2-8} 
                            & TabDDPM    & 94.28 & 0     & 94.28 & 2.86  & 0     & 14.28 \\\cline{2-8} 
                            & TLDM       & 97.14 & 0     & 97.14 & 14.28 & 0     & 0     \\\cline{2-8} 
                            & DP-CTGAN   & 14.28 & 14.28 & 2.86  & 0     & 0     & 0     \\\cline{2-8} 
                            & DP-TabDDPM & 0     & 5.72  & 0     & 0     & 0     & 1.42  \\\cline{2-8} 
                            & DP-TLDM    & 15.24 & 8.58  & 0     & 1.9   & 15.24 & 6.66  \\ \bottomrule[0.9pt]
\end{tabular}
}
\caption{Memebership inference Attack additional results containing NaiveGroundhog (NG), HistGroundhog (HG), and Closest
Distance-Hamming (CD-H), Closest Distance-L2 (CD-L) and Kernel Estimator
(KE) attacks.}
\label{tab:mia_add}
\end{table}

\subsection{Additional DP Results}

In the main section, we presented the differential privacy (DP) results for various datasets, specifically Housing, Adult, and Cardio, at a fixed noise level (\(\sigma=0.2\)) and for the Loan dataset at \(\sigma=0.5\)). To further explore the impact of DP, we now investigate how varying levels of \(\sigma\) influence model performance while maintaining a constant separation value 0.1. This analysis aims to provide a comprehensive understanding of the trade-offs between privacy and utility across different datasets and noise configurations. The detailed outcomes are presented in Figures~\ref{fig:dp_noise_housing}, \ref{fig:dp_noise_adult}, and \ref{fig:dp_noise_cardio}. 

Examination of these results reveals that our model consistently surpasses other DP-protected tabular generative models in overall performance at different noise level. Additionally, it was observed that despite maintaining a constant separation, the efficacy of our algorithm declines as the noise level, denoted by $\sigma$, is elevated.

\begin{figure}[htb]
    \centering
    \includegraphics[width=0.99\linewidth]{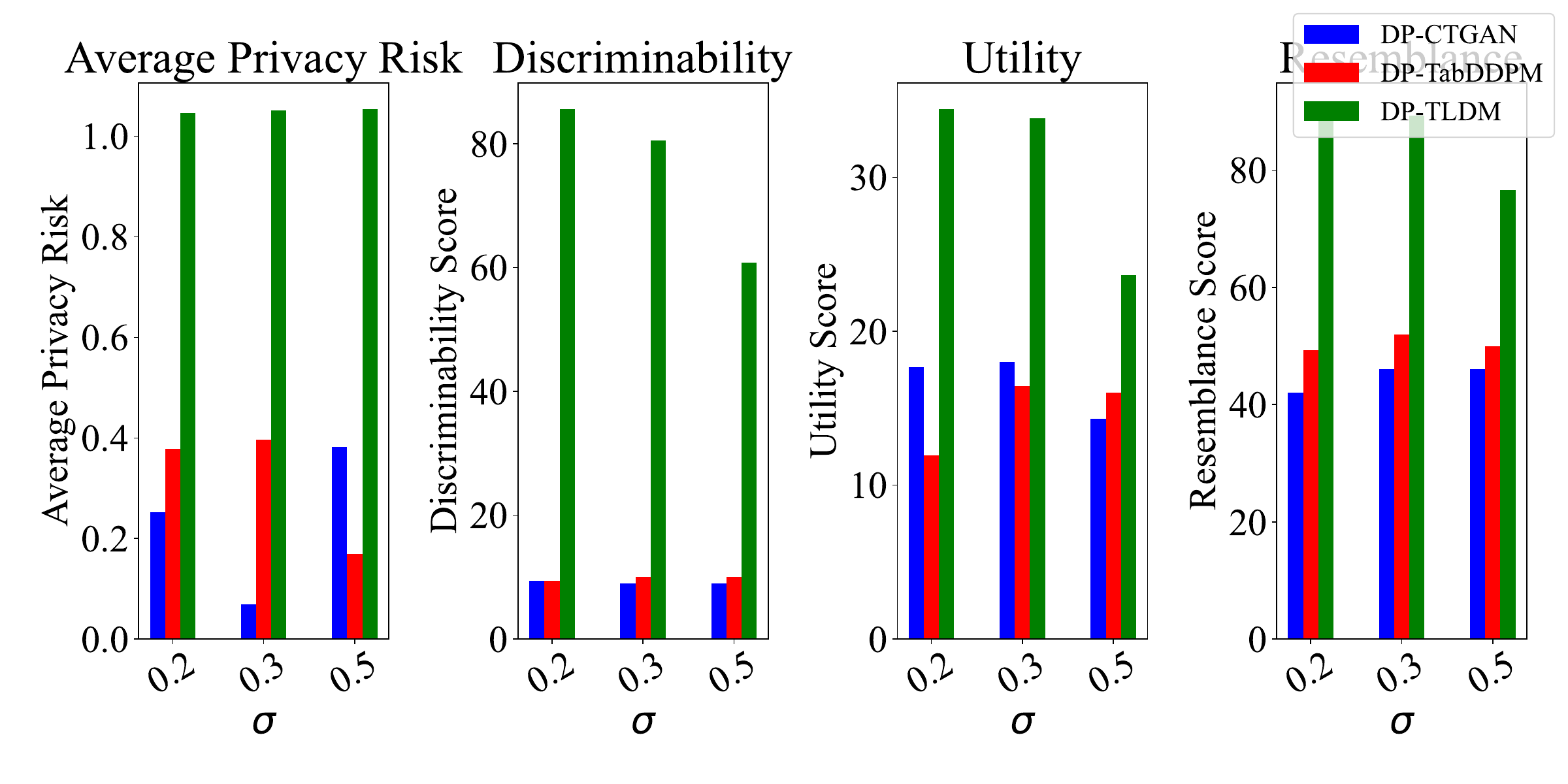}
    \caption{Impact of varying \(\sigma\) on DP for the Housing dataset with constant separation value, $sep=0.1$}
    \label{fig:dp_noise_housing}
\end{figure}

\begin{figure}[htb]
    \centering
    \includegraphics[width=0.99\linewidth]{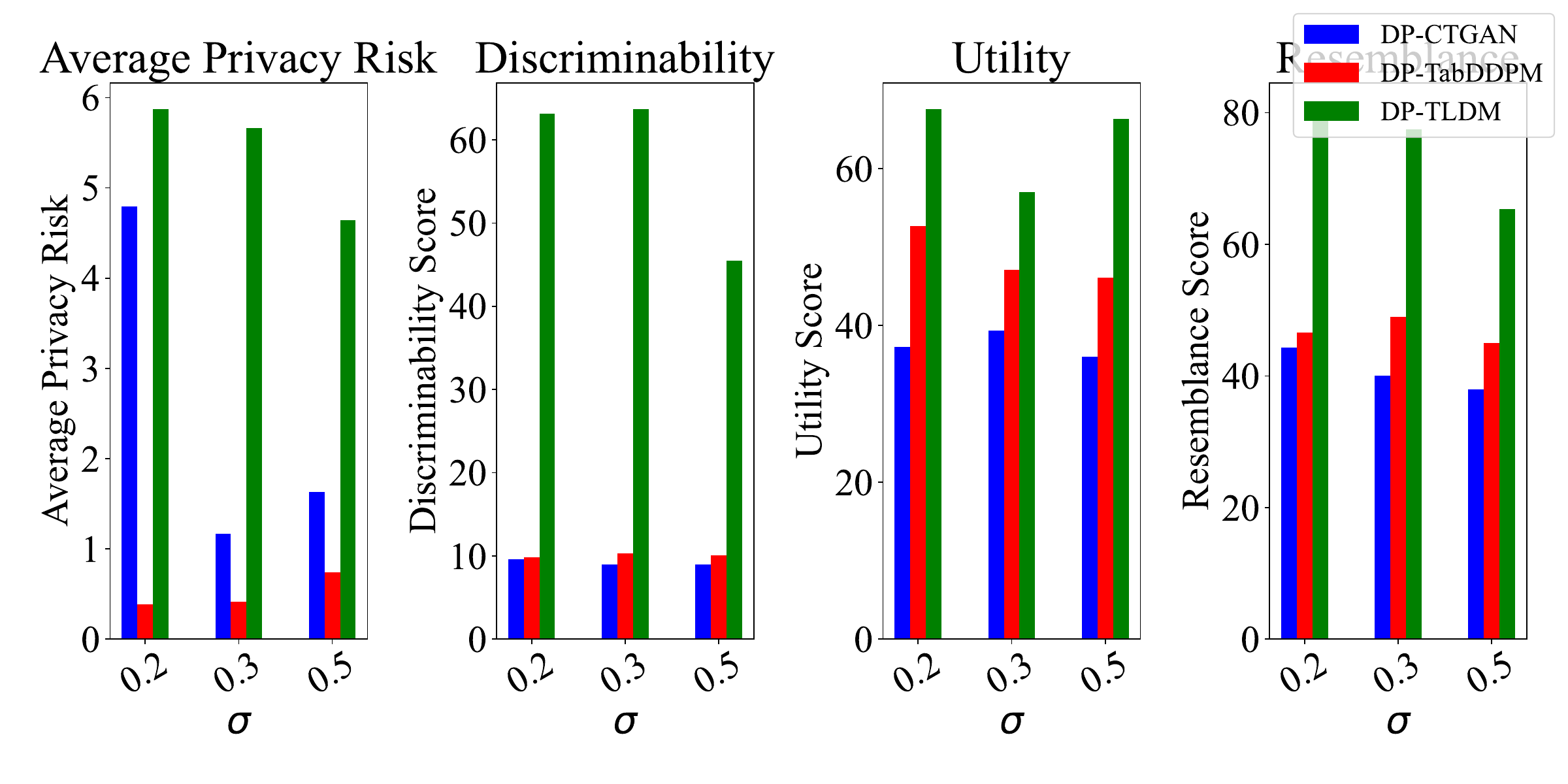}
    \caption{Impact of varying \(\sigma\) on DP for the Adult dataset with constant separation value, $sep=0.1$}
    \label{fig:dp_noise_adult}
\end{figure}

\begin{figure}[htb]
    \centering
    \includegraphics[width=0.99\linewidth]{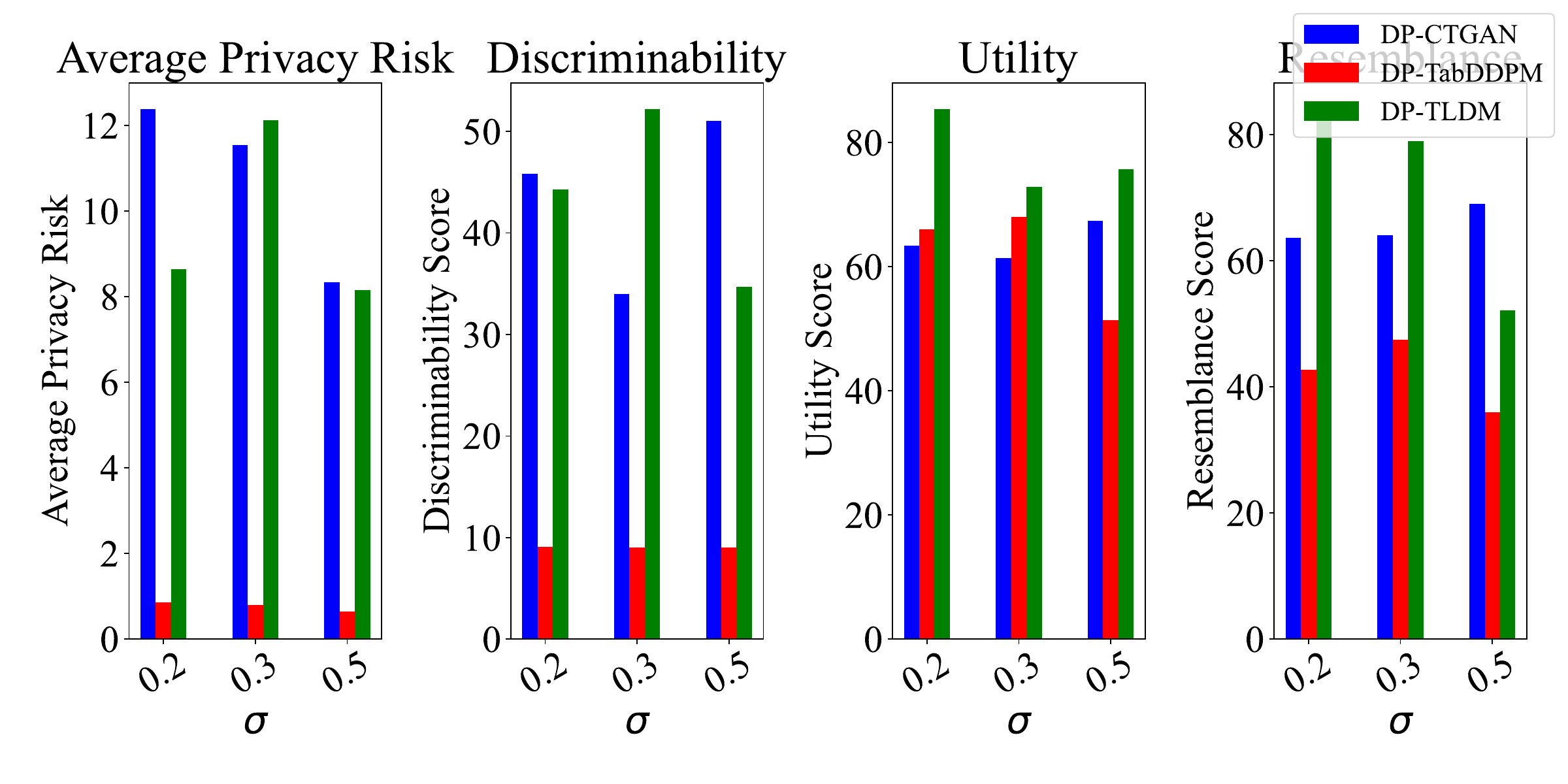}
    \caption{Impact of varying \(\sigma\) on DP for the Cardio dataset with constant separation value, $sep=0.1$}
    \label{fig:dp_noise_cardio}
\end{figure}

\end{document}